\documentclass[runningheads]{llncs}

\usepackage[mobile]{eccv}

\usepackage{eccvabbrv}

\usepackage{graphicx}
\usepackage{booktabs}

\usepackage[accsupp]{axessibility}  % Improves PDF readability for those with disabilities.

\usepackage{hyperref}

\usepackage{orcidlink}

\usepackage{array,multirow}
\usepackage{float}
\usepackage{pifont}
\newcommand{\cmark}{\ding{51}}
\newcommand{\xmark}{\ding{55}}
\usepackage{marvosym}

\begin{document}

\title{SAFNet: Selective Alignment Fusion Network for Efficient HDR Imaging}

\titlerunning{SAFNet: Selective Alignment Fusion Network for Efficient HDR Imaging}

\author{Lingtong Kong \and
%\orcidlink{0000-0003-2212-3581}
Bo Li \and
Yike Xiong \and
Hao Zhang \and
Hong Gu \and
Jinwei Chen\textsuperscript{\Letter}}

\authorrunning{L. Kong et al.}

\institute{vivo Mobile Communication Co., Ltd, China\\
\email{\{ltkong,libra,cokexiong,haozhang,guhong,jinwei.chen\}@vivo.com}}

\maketitle

\begin{abstract}
	Multi-exposure High Dynamic Range (HDR) imaging is a challenging task when facing truncated texture and complex motion. Existing deep learning-based methods have achieved great success by either following the alignment and fusion pipeline or utilizing attention mechanism. However, the large computation cost and inference delay hinder them from deploying on resource limited devices. In this paper, to achieve better efficiency, a novel Selective Alignment Fusion Network (SAFNet) for HDR imaging is proposed. After extracting pyramid features, it jointly refines valuable area masks and cross-exposure motion in selected regions with shared decoders, and then fuses high quality HDR image in an explicit way. This approach can focus the model on finding valuable regions while estimating their easily detectable and meaningful motion. For further detail enhancement, a lightweight refine module is introduced which enjoys privileges from previous optical flow, selection masks and initial prediction. Moreover, to facilitate learning on samples with large motion, a new window partition cropping method is presented during training. Experiments on public and newly developed challenging datasets show that proposed SAFNet not only exceeds previous SOTA competitors quantitatively and qualitatively, but also runs order of magnitude faster. Code and dataset is available at \href{https://github.com/ltkong218/SAFNet}{\textcolor{magenta}{https://github.com/ltkong218/SAFNet}}.
\end{abstract}

\setlength{\abovedisplayskip}{2mm}
\setlength{\belowdisplayskip}{2mm}

\section{Introduction}
\label{sec:introduction}
Human eyes are capable to perceive a broad range of illumination in natural scenes, but camera sensors suffer from limited dynamic range due to inherent hardware properties, \textit{i.e.}, the sensor's \textit{thermal noise} and \textit{full well electron capacity}~\cite{6742594}. The most common way to capture High Dynamic Range (HDR) image is to take a series of low dynamic range (LDR) photos at different exposures, and then merge them into an HDR image with increase realism~\cite{Debevec_1997}.

\begin{figure}[t]
	\centering
	\begin{subfigure}{0.49\linewidth}
		\includegraphics[width=0.97\columnwidth]{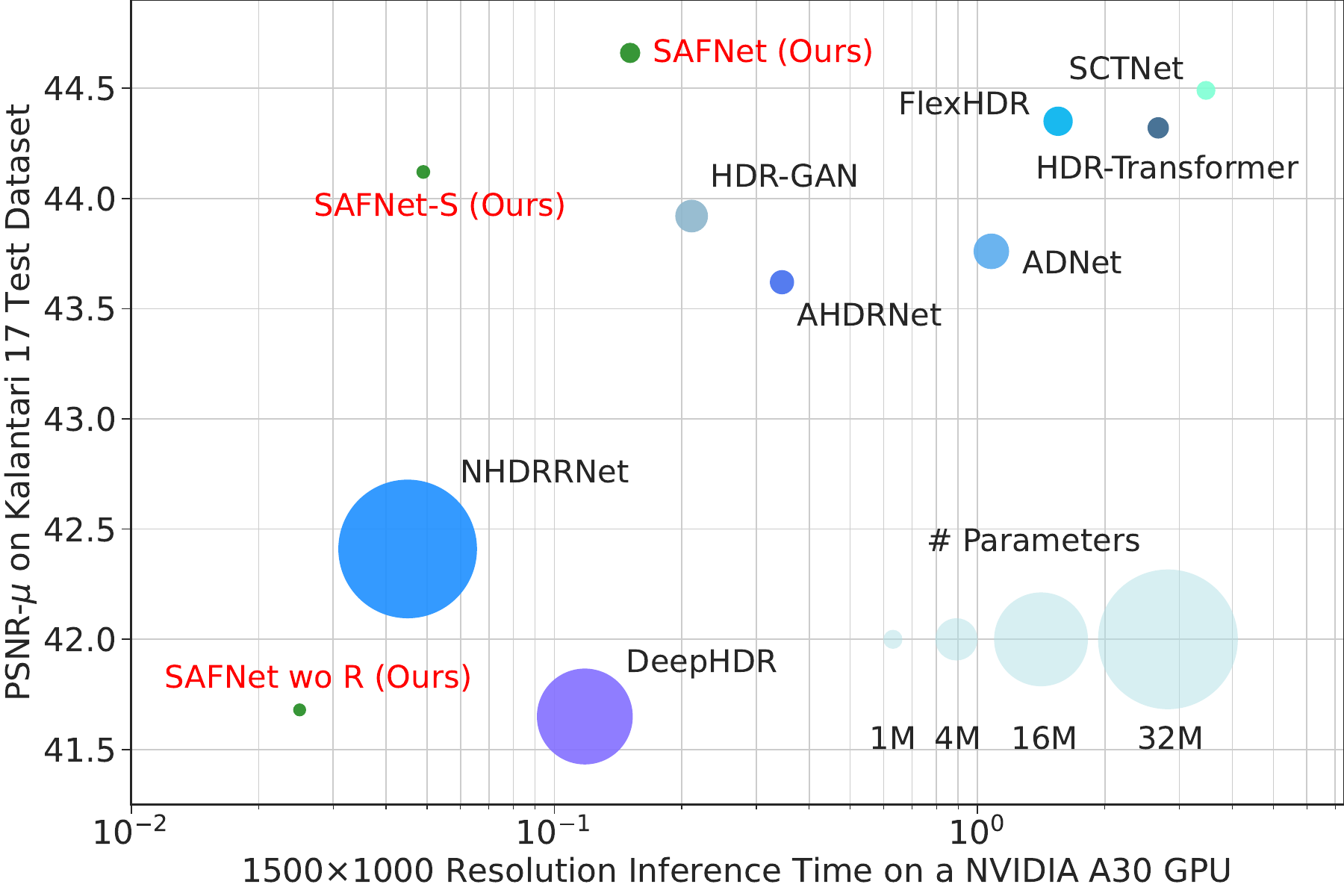}
	\end{subfigure}
	\begin{subfigure}{0.47\linewidth}
		\includegraphics[width=0.98\columnwidth]{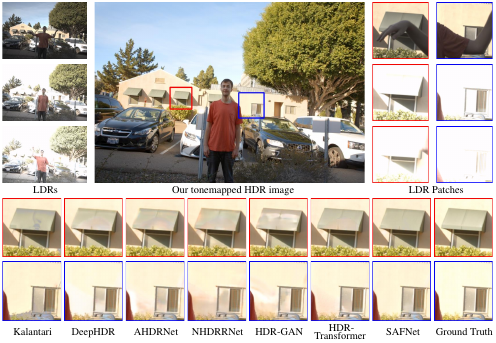}
	\end{subfigure}
	\caption{Comparison on Kalantari 17 test dataset~\cite{Kalantari_2017_ToG}. Proposed SAFNet achieves state-of-the-art HDR imaging accuracy while with fast inference speed and small model size.}
	\label{fig:1}
\end{figure}

If there is no motion or the LDR images are well aligned, existing image fusion methods can already produce faithful results~\cite{Debevec_1997,6030938,7859418,Prabhakar_2017_ICCV,8906233}. Nevertheless, dynamic objects and camera motion usually appear in shooting scenes, which results in undesirable misalignment between LDR inputs. Directly applying previous methods will yield ghosting artifacts. To deal with dynamic scenarios, traditional methods try to align the LDR input images~\cite{903475,6618998,Kang_2003_ToG} or rejecting misaligned pixels~\cite{5559003,5693088,4459868,4106952} before the fusion process. However, accurate alignment between LDR inputs with large motion and severe saturation is extremely challenging, while rejecting pixels in misaligned areas will cause insufficient content in moving regions.

As for deep learning based multi-exposure HDR imaging methods, main solutions can be summarized into two paradigms. The first class follows the alignment and fusion pipeline~\cite{Kalantari_2017_ToG,8747329,Wu_2018_ECCV,Prabhakar_2020_ECCV}, where cross-exposure motion is firstly estimated, then HDR fusion coefficients or final HDR results are generated based on aligned LDR inputs and context features. However, estimating optical flow between LDR frames under severe saturation and occlusion is error-prone. Some solutions~\cite{9881970,Yan_2023_CVPR,Chung_2023_ICCV} design special cross-exposure motion estimation networks for better alignment, but their computational complexity also increases. Differently, the second category bypasses explicit alignment, and proposes end-to-end deep networks with diverse attention mechanisms for fully spatial-/channel-wise feature interaction~\cite{Yan_2019_CVPR,9740005,8989959,Song_2022_ECCV,Liu_2022_ECCV,Tel_2023_ICCV}. There are also some methods that combine above paradigms for mutual promotion~\cite{9523096,Yan_2023_CVPR}. However, as shown in Figure~\ref{fig:1}, recent research improves HDR reconstruction accuracy by proposing increasingly complex attention mechanisms, whose large inference delay and computation cost have hindered them from deploying on power constrained devices.

To achieve better efficiency, we propose a novel Selective Alignment Fusion Network (SAFNet) for multi-frame HDR reconstruction. Different from above design concepts, we observe that not all regions in the non-reference LDR images are worthy of precise alignment. For example, if some regions in the non-reference LDR inputs are over-/under-exposed or corresponding to well-exposed texture of the reference LDR frame, these areas can be directly discarded. On the other hand, if some regions in the non-reference LDR frames contain valuable texture that is missed in the reference LDR image, accurate alignment and fusion in these regions can promote final reconstruction quality. Fortunately, motion estimation in regions with distinct texture is much easier than that are saturated~\cite{4767897,5539939}. By holding above proposition, our SAFNet performs valuable area selection and flow estimation in selected regions simultaneously, which can skip the tough yet error prone motion estimation in worthless areas, while focus the model's learning capabilities on more meaningful things. In practice, proposed SAFNet follows the successful pyramidal pipeline in optical flow networks~\cite{Ranjan_2017_CVPR,Sun_2018_CVPR,9560800}. Specifically, we use gradually refined one-channel selection probability masks to denote valuable regions during coarse-to-fine flow estimation. These masks are finally adopted to reweight fusion coefficients for flow aligned LDR inputs and generate high quality HDR result by explicit fusion operation~\cite{Kalantari_2017_ToG}. At last, for high frequency detail enhancement, a lightweight refine module is introduced at input resolution. Thanks to the valuable information in previous optical flow, selection masks and initial HDR prediction, we find that simply employing several dilated residual blocks can already achieve SOTA accuracy while with much higher efficiency. Tied to our two-stage deep network, a new window partition cropping method is presented when optimizing SAFNet, which can benefit long-distance texture aggregation and short-distance detail refinement simultaneously.

In addition to progressively advanced HDR algorithms, datasets also play an important role for evaluation. Though the dataset developed by Kalantari \textit{et al.}~\cite{Kalantari_2017_ToG} has largely facilitated the research for multi-exposure HDR imaging, only three test samples are challenging enough for visual comparison. A recent work~\cite{Tel_2023_ICCV} proposes a new HDR dataset with enriched scene motion and content. However, motion magnitude and saturation ratio in their datasets are relatively small, restricting its evaluative ability. In order to study the performance gap between different algorithms in challenging cases, we propose a new multi-exposure HDR dataset with enhanced motion range and saturated regions, that clearly distinguishes the quantitative and qualitative HDR reconstruction results among different approaches. Finally, we do experiments on the Kalantari 17 dataset~\cite{Kalantari_2017_ToG} as well as our developed Challenge123 dataset. As shown in Figure~\ref{fig:1}, the proposed SAFNet not only sets new state-of-the-art accuracy but also runs order of magnitude faster than recent Transformer-based competitors~\cite{Liu_2022_ECCV,Tel_2023_ICCV}. Main contributions of this paper can be summarized as follows:
\begin{itemize}
	\item We propose a novel SAFNet for multi-frame HDR imaging, that jointly refines valuable region masks and cross-exposure motion in selected regions, and then explicitly fuses a high quality result with much better efficiency.
	\item We provide a new challenging multi-frame HDR deghosting dataset with enhanced motion and saturation for ease of analysis.
	\item Experiments on public and newly developed datasets show that our SAFNet outperforms previous SOTA methods and runs order of magnitude faster.
\end{itemize}

\section{Related Work}
\label{sec:relatedwork}
\noindent \textbf{Traditional Methods.} Traditional multi-exposure HDR methods mainly utilize pixel rejection or motion registration techniques. The first class focuses on aligning LDR inputs globally and then discarding misaligned pixels before image fusion for deghosting. To generate error map for pixel rejection, Grosch \textit{et al.}~\cite{Grosch_2006_VMV} leverage color difference between the aligned images, Pece \textit{et al.}~\cite{5693088} employ the median threshold bitmap of the LDR inputs, Jacobs \textit{et al.}~\cite{4459868} propose weighted intensity variance analysis. Besides, Zhang \textit{et al.}~\cite{6030938} and Khan \textit{et al.}~\cite{4106952} calculate gradient-domain weight maps or probability maps of the LDR inputs, respectively. Lee \textit{et al.}~\cite{6814772} and Oh \textit{et al.}~\cite{6915885} detect moving areas by utilizing rank minimization. However, pixel rejection approach abandons useful texture in moving regions, producing unpleasing reconstruction quality.

The other motion registration-based methods rely on densely aligning the no-reference LDR inputs to the reference frame prior to merging them. Bogoni \textit{et al.}~\cite{903475} calculate optical flow as motion vectors for full image alignment. Kang \textit{et al.}~\cite{Kang_2003_ToG} transfer the LDR inputs into luminance domain according to exposure time for improving flow accuracy. Sen \textit{et al.}~\cite{Sen_2012_ToG} introduce a patch based energy minimization method that optimizes alignment and HDR fusion at the same time. Additionally, Hu \textit{et al.}~\cite{6618998} promote image alignment by propagating brightness and gradient information iteratively in a coarse-to-fine manner. However, optimizing energy function for motion estimation usually drops into local minimum. Also, their slow speed is unsuitable for real-time applications.

\noindent \textbf{Deep Learning Approaches.} Early deep learning multi-exposure HDR algorithms follow traditional alignment and fusion pipeline. Kalantari \textit{et al.}~\cite{Kalantari_2017_ToG} pioneer learning-based multi-frame HDR reconstruction by proposing a paired LDR-HDR dataset and developing a convolutional neural network (CNN) to fuse LDR inputs after flow alignment. Wu \textit{et al.}~\cite{Wu_2018_ECCV} instead adopt image-wide homography to perform background alignment, while leaving the complex foreground motions to be handled by the CNN. Despite noteworthy performance improvement over traditional methods, both \cite{Kalantari_2017_ToG} and \cite{Wu_2018_ECCV} suffer from misalignment in the presence of both large motion and severe saturation. Subsequent methods improve alignment by building more powerful cross-exposure motion estimation modules~\cite{9881970,Chung_2023_ICCV} or perform feature alignment with attention mechanism~\cite{9523096,Yan_2023_CVPR}. However, the large computation cost and running time hinders their development on mobile devices.

Yan \textit{et al.} address some limitations of the predecessors by introducing a spatial attention module~\cite{Yan_2019_CVPR}, and further constructing a non-local block to improve global consistency~\cite{8989959}. Niu \textit{et al.}~\cite{9387148} leverage Generative Adversarial Network (GAN) to synthesize realistic content which is missing in the LDR inputs. Furthermore, Xiong \textit{et al.}~\cite{Xiong_2021_ACMMM} decompose HDR imaging into ghost-free image fusion and ghost-based image restoration. Ye \textit{et al.}~\cite{Ye_2021_ACMMM} propose a progressive feature fusion network that compares and selects appropriate LDR regions to generate high quality result. Above attention-based and region selective HDR algorithms can facilitate deghosting in fusion stage. However, their results fall behind current state-of-the-arts due to the limited long-range texture aggregation ability.

Recently, transformers have shown better ability to capture long-range dependency than CNN due to their multi-head self-attention mechanism. Song \textit{et al.}~\cite{Song_2022_ECCV} separate LDR inputs into ghost and non-ghost regions, and then selectively apply either transformer or CNN to perform HDR reconstruction. Liu \textit{et al.}~\cite{Liu_2022_ECCV} integrate vision transformer with convolution to explore both local and global relationship and obtain remarkable results. Furthermore, Yan \textit{et al.}~\cite{Yan_2023_CVPR} propose a HyHDRNet consisting of a content alignment subnetwork and a transformer-based fusion subnetwork for performance improvement. Tel \textit{et al.}~\cite{Tel_2023_ICCV} introduce a SCTNet which integrates both spatial and channel attention modules into the transformer-based network to enhance semantic consistency. Nevertheless, transformer-based methods suffer from large inference delay. Moreover, their patch-based prediction manner is unable to aggregate cross-patch texture produced by large motion, which is common in high resolution imagery.

\begin{figure*}[t]
	\centering
	\includegraphics[width=0.98\textwidth]{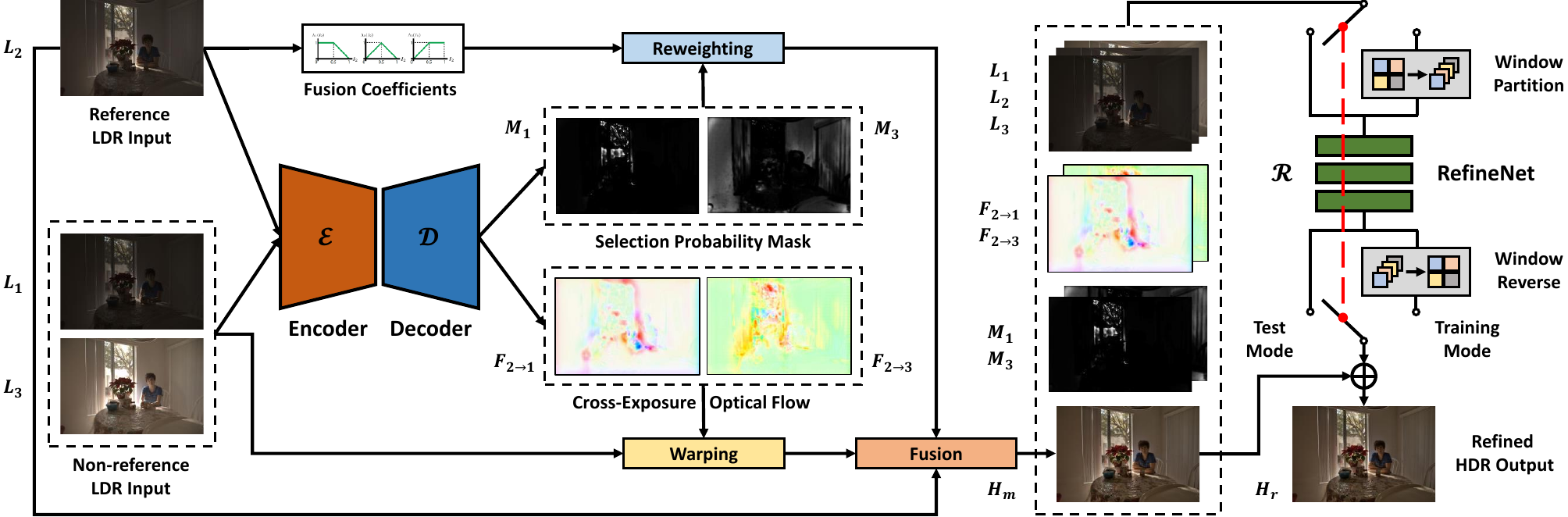}
	\caption{Overall architecture of our SAFNet. It contains a pyramid encoder, a coarse-to-fine decoder, and a refinement subnetwork. The linked switch selects path including window partition and window reverse during training, while skip them in evaluation.}
	\label{fig:2}
\end{figure*}

\section{Proposed Method}
\label{sec:proposedmethod}
\noindent \textbf{Overview.} Given three LDR images $L_1, L_2, L_3$ from a dynamic scene with different exposures as input, our goal is to generate a ghost-free HDR image $H_r$ with consistent scene structure as the reference image $L_2$. Figure~\ref{fig:2} depicts overall architecture of the proposed SAFNet, containing three subnetworks, \textit{i.e.}, the pyramid encoder $\mathcal{E}$, the coarse-to-fine decoder $\mathcal{D}$, and the detail refine module $\mathcal{R}$. SAFNet first performs an extraction phase to retrieve a pyramid of features from each input frame $L_i$ by the encoder $\mathcal{E}$. Then, it jointly refines selection probability masks $M_1, M_3$ together with cross-exposure optical flow $F_{2\rightarrow1}, F_{2\rightarrow3}$ in selected regions by the decoder $\mathcal{D}$. Furthermore, a high quality HDR image $H_m$ is explicitly merged by flow aligned LDR inputs and selection masks reweighted fusion coefficients. Finally, our SAFNet generates a refined HDR image $H_r$ by the refine network $\mathcal{R}$ based on LDR inputs $L_i$, cross-exposure motion $F_{2\rightarrow i}$, selection probability masks $M_i$ and merged HDR image $H_m$. In addition to network architecture, training loss function and developed window partition cropping method are also elaborated in this section.

\noindent \textbf{Pyramid Encoder.} Like previous methods~\cite{Kalantari_2017_ToG,Yan_2019_CVPR,Yan_2023_CVPR,Tel_2023_ICCV}, we first map the LDR frames $L_i$ to the HDR linear domain by using gamma correction as follows:
\begin{equation}
	H_i = L_i^{\gamma} / t_i, \quad i=1,2,3,
\end{equation}
where $t_i$ denotes the exposure time of LDR image $L_i$, $\gamma$ is the gamma parameter, which is set to $2.2$. By concatenating $L_i$ and $H_i$ along the channel dimension, we obtain three 6-channel tensors $X_i = [L_i, H_i]$ as network input.

Inspired by the success of pyramidal flow estimation architectures~\cite{Ranjan_2017_CVPR,Sun_2018_CVPR,Kong_2020_ICIP,9560800}, our encoder network $\mathcal{E}$ extracts multi-scale pyramid features to better estimate cross-exposure optical flow in the challenging large motion and heavy saturation cases. Purposely, the encoder is built by a block of two 3$\times$3 convolutions in each pyramid level, with strides 2 and 1, respectively. The parameter shared encoder extracts 4 levels of pyramid features, counting 8 convolution layers, each followed by a PReLU activation~\cite{7410480}. With gradually decimated spatial size, it keeps the feature channels to 40 among all 4 scales, generating pyramid features $\phi_{1}^{k}, \phi_{2}^{k}, \phi_{3}^{k}$ in level $k$ ($k=1,2,3,4$) for LDR inputs $L_1, L_2, L_3$, separately.

\noindent \textbf{Coarse-to-Fine Decoder.} To deal with the large displacement challenge for motion estimation, we follow the successful pyramid optical flow networks~\cite{Ranjan_2017_CVPR,Sun_2018_CVPR,9560800}, that adopt coarse-to-fine warping strategy and predicts easier residual flow at each scale. After extracting meaningful pyramid features, the decoder $\mathcal{D}^{k}$ iteratively refines cross-exposure optical flow by backward warping pyramid features $\phi_{1}^{k}, \phi_{3}^{k}$ to generate $\tilde{\phi}_{1}^{k}, \tilde{\phi}_{3}^{k}$ according to $F_{2\rightarrow1}^{k}$ and $F_{2\rightarrow3}^{k}$, respectively, where $\mathcal{D}^{k}$ means this parameter shared decoder $\mathcal{D}$ is called in level $k$. However, unlike traditional optical flow task, cross-exposure motion estimation is much more challenging due to co-existence of large motion and severe saturation. Instead of designing complex flow estimation network for overall improvement as previous~\cite{9881970,Yan_2023_CVPR,Chung_2023_ICCV}, we observe that not all regions of $F_{2\rightarrow1}, F_{2\rightarrow3}$ are worthy of accurate prediction. Therefore, besides cross-exposure optical flow $F_{2\rightarrow1}^{k-1}, F_{2\rightarrow3}^{k-1}$, the decoder network $\mathcal{D}^{k}$ further predicts two selection probability masks $M_1^{k-1}, M_3^{k-1}$ to denote valuable regions of $F_{2\rightarrow1}^{k-1}, F_{2\rightarrow3}^{k-1}$ during coarse-to-fine flow estimation, respectively. Specifically, $M_1^{k-1}$ and $M_3^{k-1}$ are one-channel tensors exported by sigmoid function whose elements range from 0 to 1. Different from previous cascading HDR reconstruction pipeline~\cite{Kalantari_2017_ToG}, \textit{i.e.}, first optical flow then fusion, our jointly refined $M_1, M_3$ and $F_{2\rightarrow1}, F_{2\rightarrow3}$ can benefit each other. First, $M_1, M_3$ can inform the decoder to focus on estimating $F_{2\rightarrow1}, F_{2\rightarrow3}$ in the identified areas. In turn, better estimated $F_{2\rightarrow1}, F_{2\rightarrow3}$ can aggregate valuable pyramid features from non-reference frames to promote further region identification and residual flow estimation. Therefore, accuracy and efficiency are both improved. In summary, input and output of the coarse-to-fine decoder can be formulated as:
\begin{equation}
		[F_{2\rightarrow1}^{k-1}, F_{2\rightarrow3}^{k-1}, M_1^{k-1}, M_3^{k-1}] = \mathcal{D}^{k}([F_{2\rightarrow1}^{k}, F_{2\rightarrow3}^{k}, M_1^{k}, M_3^{k}, \tilde{\phi}_{1}^{k}, \phi_{2}^{k}, \tilde{\phi}_{3}^{k}]),
\end{equation}
where $\mathcal{D}^{k}$ ($k=1,2,3,4$) denotes the iterative refinement in level $k$, $[\cdot]$ means feature concatenation. The initial value of $F_{2\rightarrow1}^{4}, F_{2\rightarrow3}^{4}, M_1^{4}, M_3^{4}$ are all set to 0, while the final prediction by $\mathcal{D}^{1}$ are written as $F_{2\rightarrow1}, F_{2\rightarrow3}, M_1, M_3$.

\begin{figure}[t]
	\centering
	\includegraphics[width=0.88\columnwidth]{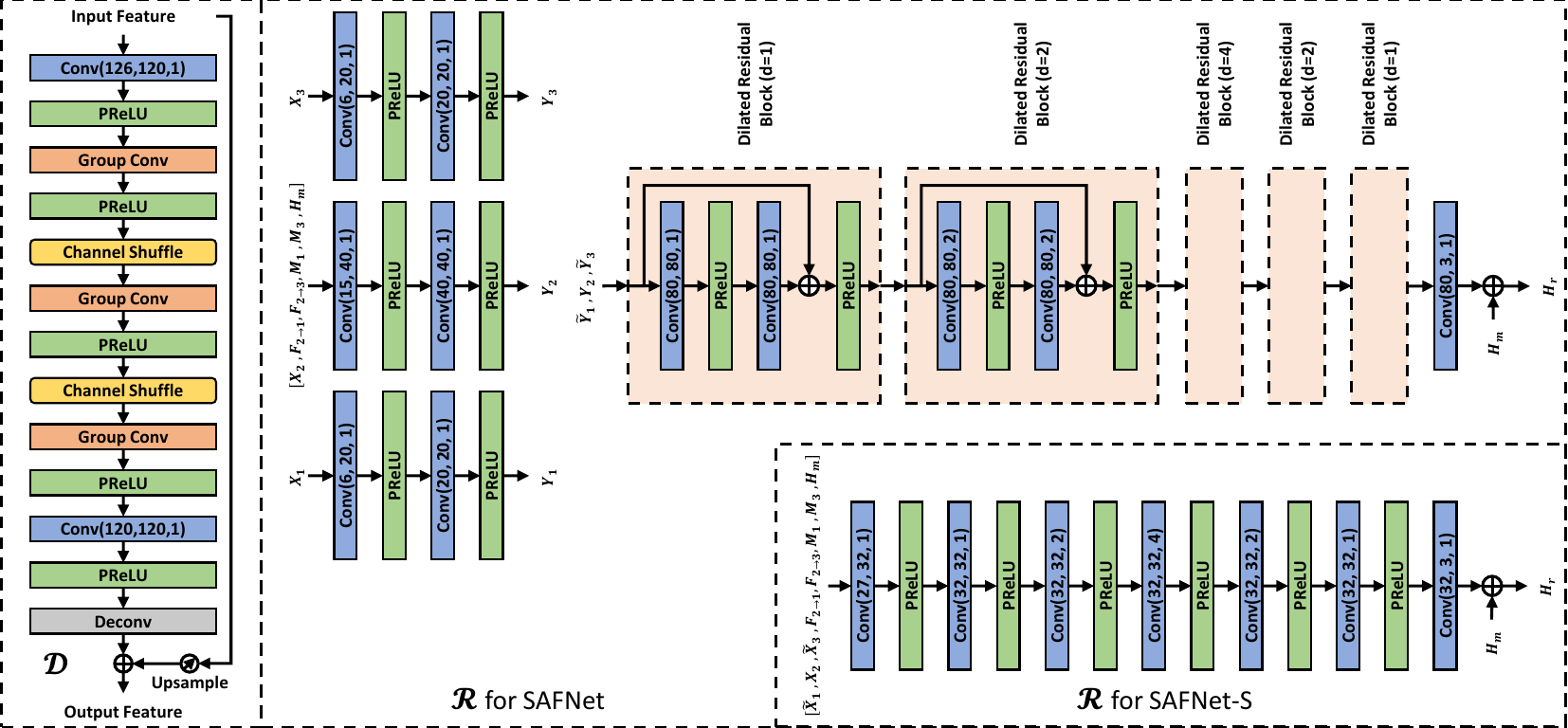}
	\caption{Details of the decoder $\mathcal{D}$ and the refine network $\mathcal{R}$ for SAFNet and SAFNet-S. Arguments of `Conv' from left to right are input channels, output channels and dilation. All convolutions have 3$\times$3 kernel size. Stride is equal to dilation for each `Conv'.}
	\label{fig:3}
\end{figure}

Concretely, the decoder $\mathcal{D}$ is consist of a block of five 3$\times$3 convolutions and one 4$\times$4 deconvolution, with strides 1 and 1/2, respectively. A PReLU~\cite{7410480} follows each convolution layer. Feature channels of intermediate layers of $\mathcal{D}$ are all set to 120. Following the success of efficient flow estimation in \cite{9560800}, the middle three convolutions of $\mathcal{D}$ are group convolution with group number equal to 3, which are separated by channel shuffle operation. Figure~\ref{fig:3} shows structure details of the decoder that is shared among all pyramid levels.

\begin{figure}[t]
	\centering
	\includegraphics[width=0.5\columnwidth]{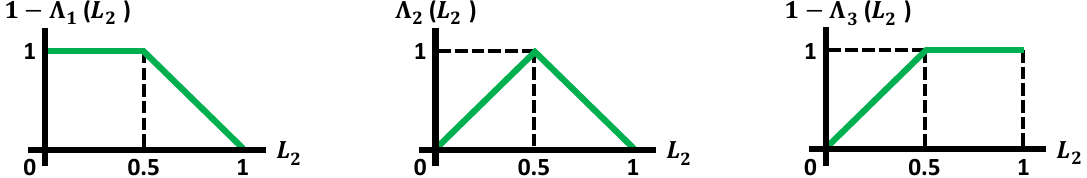}
	\caption{Functions to define initial HDR fusion coefficients.}
	\label{fig:4}
\end{figure}

\noindent \textbf{Explicit HDR Fusion.} The final goal of predicted optical flow $F_{2\rightarrow1}, F_{2\rightarrow3}$ and selection probability masks $M_1, M_3$ is to merge a high quality HDR image $H_m$ as close as possible to the ground truth $H_{gt}$. As shown in Figure~\ref{fig:2}, there are two preliminary steps before the HDR fusion procedure. At first, we use estimated optical flow $F_{2\rightarrow1}, F_{2\rightarrow3}$ to align the non-reference linear domain images $H_1, H_3$, and generate warped images of $\tilde{H}_{1}, \tilde{H}_{3}$ as follows:
\begin{equation}
	\tilde{H}_{1} = w(H_{1}, F_{2\rightarrow1}), \; \tilde{H}_{3} = w(H_{3}, F_{2\rightarrow3}),
\end{equation}
where $w$ means backward warping~\cite{Ranjan_2017_CVPR,Sun_2018_CVPR}. Secondly, predicted selection probability masks $M_1, M_3$ are employed to reweight initial fusion coefficients for ghost-suppressed HDR synthesis. Considering that the predicted optical flow $F_{2\rightarrow1}, F_{2\rightarrow3}$ by SAFNet are relatively accurate only in regions where $M_1, M_3$ contain a relatively large selection probability. Therefore, to eliminate ghosting artifacts when fusing unrelated textures, we multiply the initial fusion coefficients of $\tilde{H}_{1}, \tilde{H}_{3}$ with their selection probability masks $M_1, M_3$, respectively. Meanwhile, the unselected parts of initial fusion coefficients of $\tilde{H}_{1}, \tilde{H}_{3}$ are transferred to the reference image $H_2$ for normalization. Formulaically, proposed reweighted fusion coefficients can be computed by:
\begin{equation}
	\begin{split}
		& W_1 = \Lambda_1 \odot M_1, \quad W_3 = \Lambda_3 \odot M_3, \\
		W_2 = & \; \Lambda_2 + \Lambda_1 \odot (1 - M_1) + \Lambda_3 \odot (1- M_3),
	\end{split}
\end{equation}
where $\Lambda_1, \Lambda_2, \Lambda_3$ are initial HDR fusion coefficients for inputs $H_1, H_2, H_3$, that are defined in Figure~\ref{fig:4}. $W_1, W_2, W_3$ are the reweighted fusion coefficients of SAFNet. $\odot$ means element-wise multiplication. Given optical flow aligned input images and selection masks reweighted fusion coefficients, we can merge a high quality HDR image explicitly by:
\begin{equation}
	H_m = W_1 \odot \tilde{H}_{1} + W_2 \odot H_2 + W_3 \odot \tilde{H}_{3}.
	\label{eq:5}
\end{equation}

\noindent \textbf{Refine Network.} Explicit fusion approach in Eq.~\ref{eq:5} can already fuse a high quality HDR image. However, it can not synthesize textures that are truncated or occluded in all input frames. To compensate for missing content and remove potential ghost, a refine network $\mathcal{R}$ is introduced in Figure~\ref{fig:2}, whose details are shown in Figure~\ref{fig:3}. Specifically, $\mathcal{R}$ is a fully convolutional network that works at original input resolution to enhance high frequency details. First, three independent feature extractors, each including two 3$\times$3 convolutions, are adopted to extract local features $Y_1, Y_2$ and $Y_3$ from inputs $X_1$, $[X_2, F_{2\rightarrow1}, F_{2\rightarrow3}, M_1, M_3, H_m]$ and $X_3$, respectively. Then, aligned shallow features of $\tilde{Y}_1, \tilde{Y}_3$ are obtained by backward warping $Y_1, Y_3$ according to flow $F_{2\rightarrow1}, F_{2\rightarrow3}$, separately. Finally, the concatenated feature of $[\tilde{Y}_1, Y_2, \tilde{Y}_3]$ is forwarded to an aggregation module, containing five dilated residual blocks and one convolution, to estimate residual details and yield a refined HDR image $H_r$ as our final prediction.

\noindent \textbf{Loss Function.} Since HDR images are typically viewed after tonemapping, we use $\mu$-law function to map the image from linear domain to the tonemapped domain as follows:
\begin{equation}
	T(H) = log(1 + \mu H) / log(1 + \mu), \quad \mu = 5000,
\end{equation}
where $H$ denotes the HDR image in linear domain, $\mu$ is the compression parameter. Following methods~\cite{Liu_2022_ECCV,Yan_2023_CVPR,Tel_2023_ICCV}, we employ the weighted $\mathcal{L}_1$ loss and perceptual loss $\mathcal{L}_p$~\cite{Johnson_2016_ECCV} between our refined output $H_r$ and the ground truth $H_{gt}$ to supervise HDR reconstruction of SAFNet by:
\begin{equation}
	\mathcal{L}_r = \mathcal{L}_1(T(H_r), T(H_{gt})) + \alpha \mathcal{L}_p(T(H_r), T(H_{gt})),
\end{equation}
where $\mathcal{L}_p$ measures the distance on multi-scale features extracted by a pre-trained VGG-16 network, $\alpha$ is the weighting parameter set to $0.01$. Additionally, we add an auxiliary brightness reconstruction loss $\mathcal{L}_m$ to guide the learning of alignment and fusion for the merged HDR image $H_m$ as:
\begin{equation}
	\mathcal{L}_m = \mathcal{L}_1(T(H_m), T(H_{gt})) + \mathcal{L}_{c}(T(H_m), T(H_{gt})),
\end{equation}
where $\mathcal{L}_{c}$ is the census loss~\cite{Simon_2018_AAAI,Kong_2022_CVPR,Kong_2022_TCSVT} calculating the soft Hamming distance between census-transformed image patches of size 7$\times$7. In summary, our total training loss is the combination of $\mathcal{L}_r$ and $\mathcal{L}_m$, that can be expressed as:
\begin{equation}
	\mathcal{L} = \mathcal{L}_r + \beta \mathcal{L}_m,
\end{equation}
where $\beta$ is the trade-off coefficient, which is set to $0.1$.

\noindent \textbf{Window Partition Cropping.} Recent multi-frame HDR methods~\cite{Liu_2022_ECCV,Yan_2023_CVPR,Tel_2023_ICCV} crop 128$\times$128 image patches during training, that can generate enough data with diverse saturation and occlusion for sufficient learning. However, their relatively small crop size will block long-range texture aggregation in large motion cases. To deal with this problem, we propose a new window partition cropping method during optimization, that is bound to our two-stage SAFNet. As shown in Figure~\ref{fig:2}, the merged $H_m$ is generated on large patches of 512$\times$512 to better aggregate long-range inter-frame textures. On the other hand, the refined $H_r$ is predicted on small patches of 128$\times$128 to better synthesize local details. We unify above two different cropping sizes by window partition and reverse operations, where the additional size is first shifted to the batch dimension and then shifted back. In test stage, window partition and reverse are discarded.

\section{Proposed Dataset}
\label{sec:proposeddataset}

\begin{figure*}[t]
	\centering
	\begin{minipage}{0.48\linewidth}
		\begin{figure}[H]
			\small
			\centering
			\begin{tabular}{c @{\hskip 0.01in} c @{\hskip 0.01in} c}
				\includegraphics[width=0.32\linewidth]{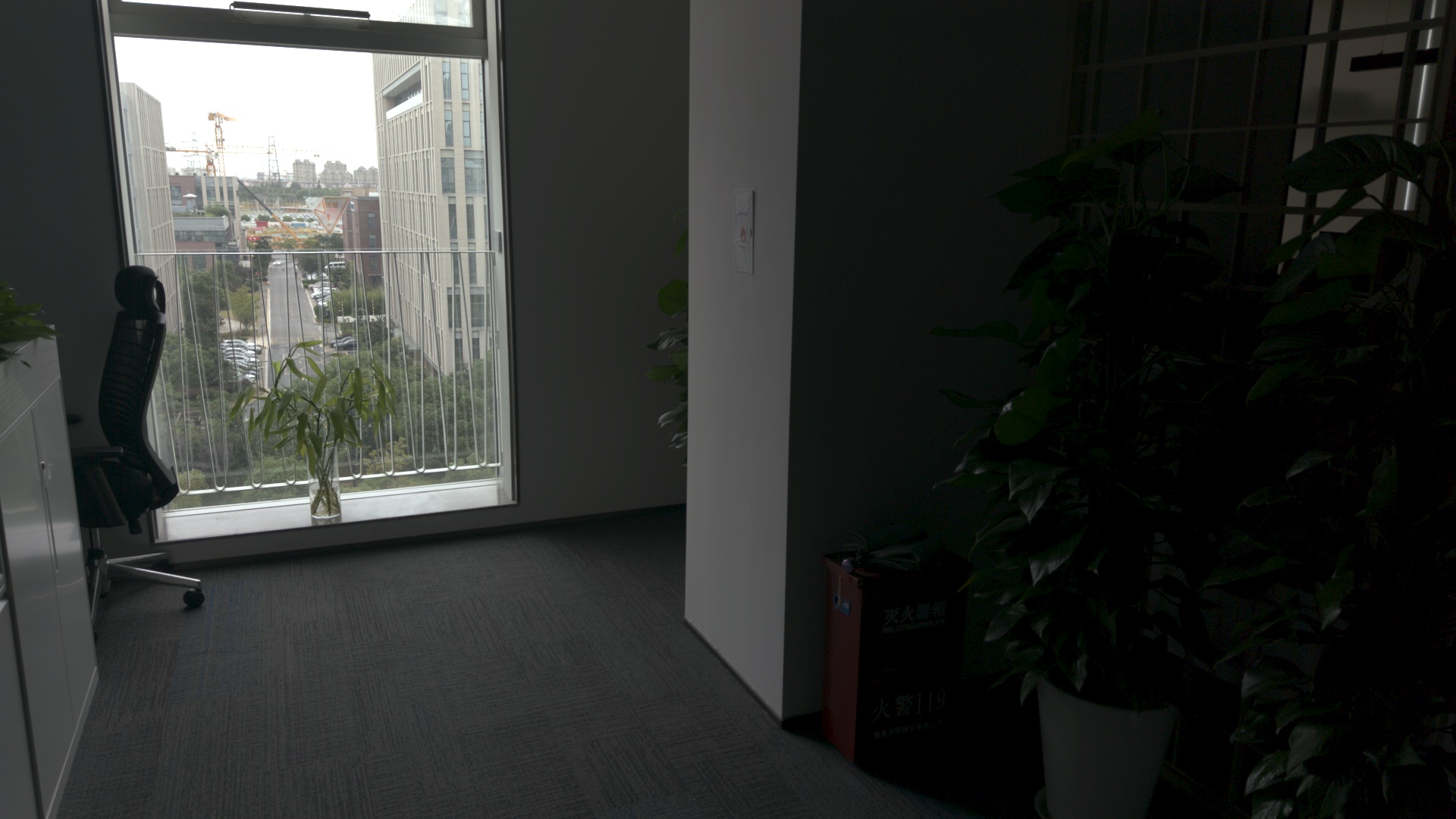}
				&
				\includegraphics[width=0.32\linewidth]{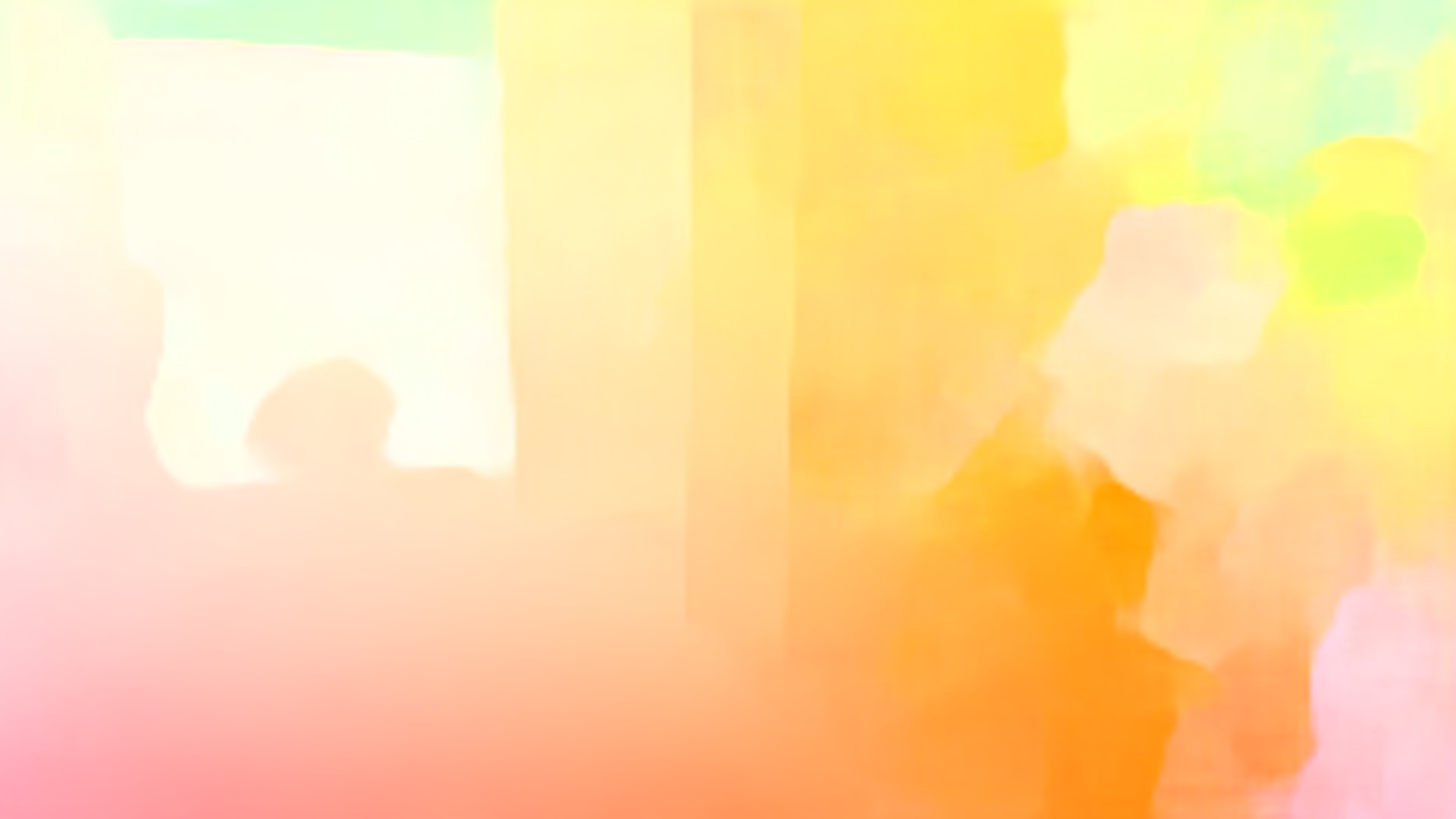}
				&
				\includegraphics[width=0.32\linewidth]{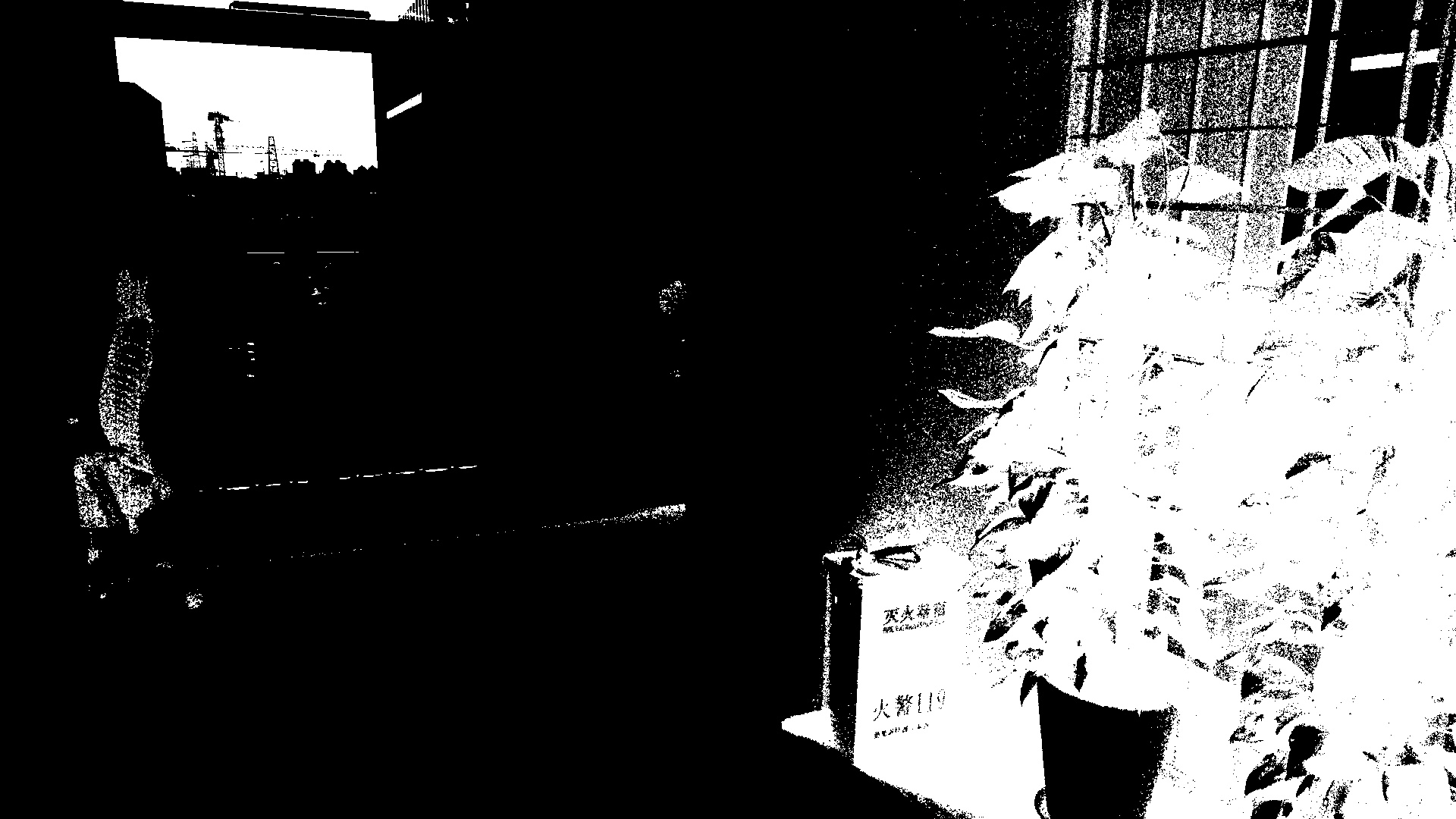}
				\\
			\end{tabular}
			\caption{Visual example of LDR input $L_2$, optical flow from $L_2$ to $L_1$ and saturated regions of $L_2$ in proposed dataset.}
			\label{fig:5}
		\end{figure}
	\end{minipage}
	\hspace{0.00\linewidth}
	\begin{minipage}{0.48\linewidth}
		\begin{table}[H]
			\renewcommand{\arraystretch}{1.0}
			\tabcolsep=0.6mm
			{\scriptsize
				\centering
				\caption{Statistics comparison among different multi-exposure HDR datasets.}
				\begin{tabular}{l|ccc}
					\toprule
					Statistics (Avg) & Kalantari~\cite{Kalantari_2017_ToG} & Tel~\cite{Tel_2023_ICCV} & Ours \\
					\midrule
					Motion Magnitude & 20.1 & 16.2 & 128.7 \\
					Saturation Ratio & 0.061 & 0.073 & 0.201 \\
					\bottomrule
				\end{tabular}
				\label{tab:1}}
		\end{table}
	\end{minipage}
\end{figure*}

The existing labeled multi-exposure HDR datasets~\cite{Froehlich_2014,Kalantari_2017_ToG,Tel_2023_ICCV} have facilitated research in related fields. However, results of recent methods~\cite{Liu_2022_ECCV,9881970,Yan_2023_CVPR,Tel_2023_ICCV} tend to be saturated due to their limited evaluative ability~\cite{Kalantari_2017_ToG,Tel_2023_ICCV}. We attribute this phenomenon to most of their samples having relatively small motion magnitude between LDR inputs and relatively small saturation ratio of the reference image. To probe the performance gap between different algorithms, we propose a new challenging multi-exposure HDR dataset with enhanced motion range and saturated regions. Our proposed Challenge123 dataset follows the same collection pipeline as \cite{Kalantari_2017_ToG}, but use a vivo X90 Pro+ phone equipped with Sony IMX 989 sensor. There are 96 training samples and 27 test samples in our dataset, whose construction details can be found in our supplementary material.

To highlight the differences of our dataset over existing ones, we use a pre-trained flow network~\cite{9560800} to estimate optical flow from $L_2$ to $L_1$ and calculate saturated area ratio of $L_2$ for every test sample in each dataset. Figure~\ref{fig:5} depicts one visual example of this statistic approach. By averaging motion magnitude and saturation ratio over all test samples in each HDR dataset, we summarize the statistics comparison of different datasets in Table~\ref{tab:1}. As can be seen, motion magnitude and saturation ratio of proposed Challenge123 dataset are about 7$\times$ and 3$\times$ larger than existing HDR datasets~\cite{Kalantari_2017_ToG,Tel_2023_ICCV}, respectively. It is worth noting that our dataset is complementary to existing ones~\cite{Kalantari_2017_ToG,Tel_2023_ICCV}, which aims to widen the performance gap between different algorithms for ease of analysis, even if it is less natural than them. For example, the challenging regions of~\cite{Kalantari_2017_ToG} where the man is waving his arms contain more than 150 pixels displacement, which is similar as the average motion magnitude of our dataset.

\begin{table*}[t]
	\renewcommand{\arraystretch}{0.85}
	{\scriptsize
		\centering
		\setlength\tabcolsep{0.15pt}
		\caption{Quantitative comparison with methods on Kalantari 17 test dataset~\cite{Kalantari_2017_ToG}. For each item, the best result is \textcolor{red}{\textbf{boldfaced}}, and the second best is \textcolor{blue}{\underline{underlined}}. Time and FLOPs are measured on one NVIDIA A30 GPU under 1500$\times$1000 resolution.}
		\begin{tabular}{l|ccccc|ccc}
			\toprule
			Method & PSNR-$\mu$ & PSNR-$l$ & SSIM-$\mu$ & SSIM-$l$ & HDR-VDP2 & Time (s) & Params (M) & FLOPs (T) \\
			\midrule
			Sen~\cite{Sen_2012_ToG} & 40.80 & 38.11 & 0.9808 & 0.9721 & 59.38 & - & - & - \\
			Kalantari~\cite{Kalantari_2017_ToG} & 42.67 & 41.23 & 0.9888 & 0.9846 & 65.05 & - & - & - \\
			DeepHDR~\cite{Wu_2018_ECCV} & 41.65 & 40.88 & 0.9860 & 0.9858 & 64.90 & 0.118 & 16.61 & 1.304 \\
			AHDRNet~\cite{Yan_2019_CVPR} & 43.63 & 41.14 & 0.9900 & 0.9702 & 64.61 & 0.345 & 1.52 & 2.170 \\
			NHDRRNet~\cite{8989959} & 42.41 & 41.43 & 0.9877 & 0.9857 & 61.21 & \textcolor{red}{\bf 0.045} & 31.56 & \textcolor{blue}{\underline{0.421}} \\
			HDR-GAN~\cite{9387148} & 43.92 & 41.57 & 0.9905 & 0.9865 & 65.45 & 0.211 & 2.56 & 1.455 \\
			ADNet~\cite{9523096} & 43.87 & 41.69 & 0.9925 & 0.9885 & 65.56 & 1.078 & 2.96 & 4.648 \\
			FlexHDR~\cite{9881970} & 44.35 & \textcolor{blue}{\underline{42.60}} & \textcolor{blue}{\underline{0.9931}} & 0.9902 & 66.56 & 1.550 & 2.12 & - \\
			HDR-Transformer~\cite{Liu_2022_ECCV} & 44.32 & 42.18 & 0.9916 & 0.9884  & 66.03 & 2.673 & 1.22 & 1.886 \\
			HyHDRNet~\cite{Yan_2023_CVPR} & \textcolor{blue}{\underline{44.64}} & 42.47 & 0.9915 & 0.9894 & 66.05 & - & - & - \\
			SCTNet~\cite{Tel_2023_ICCV} & 44.49 & 42.29 & 0.9924 & 0.9887 & 66.65 & 3.466 & \textcolor{blue}{\underline{0.99}} & 1.547 \\
			SAFNet-S (Ours) & 44.12 & 42.50 & 0.9928 & \textcolor{blue}{\underline{0.9907}} & \textcolor{blue}{\underline{66.75}} & \textcolor{blue}{\underline{0.049}} & \textcolor{red}{\bf 0.57} & \textcolor{red}{\bf 0.146} \\
			SAFNet (Ours) & \textcolor{red}{\bf 44.66} & \textcolor{red}{\bf 43.18} & \textcolor{red}{\bf 0.9932} & \textcolor{red}{\bf 0.9917} & \textcolor{red}{\bf 66.93} & 0.151 & 1.12 & 0.976 \\
			\bottomrule
		\end{tabular}
		\label{tab:2}}
\end{table*}

\section{Experiments}
\label{sec:experiments}

\begin{figure}[t]
	\centering
	\begin{subfigure}{0.48\linewidth}
		\includegraphics[width=1.00\columnwidth]{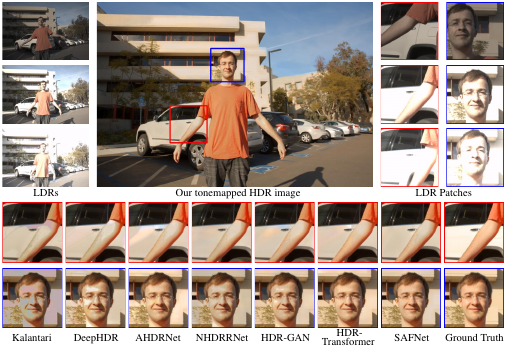}
	\end{subfigure}
	\begin{subfigure}{0.50\linewidth}
		\includegraphics[width=1.00\columnwidth]{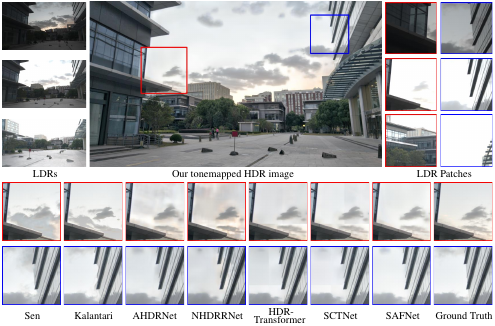}
	\end{subfigure}
	\caption{Visual comparison on Kalantari~\cite{Kalantari_2017_ToG} (left) and Challenge123 (right) test sets.}
	\label{fig:6}
\end{figure}

\noindent \textbf{Datasets.} We first evaluate proposed SAFNet on the conventional Kalantari 17 dataset~\cite{Kalantari_2017_ToG} including 74 training samples and 15 test samples. Then, we compared our algorithm with recent SOTA methods on the proposed Challenge123 dataset consist of 96 training scenes and 27 test scenes. Finally, we analyze the generalization ability of diverse approaches on the Sen's dataset~\cite{Sen_2012_ToG}.

\noindent \textbf{Evaluation Metrics.} For quantitative comparison, we calculate five commonly used metrics, \textit{i.e.}, PSNR-$\mu$, PSNR-$l$, SSIM-$\mu$, SSIM-$l$ and HDR-VDP2~\cite{Mantiuk_2011_ToG}, where -$\mu$ denotes tonemapped domain and -$l$ means linear domain.

\noindent \textbf{Implementation Details.} We implement proposed method in PyTorch, and use Kalantari 17~\cite{Kalantari_2017_ToG} or our Challenge123 training set to train SAFNet from scratch for separate comparison. Proposed network is optimized by Adam algorithm with $\beta_1$=$0.9$, $\beta_2$=$0.999$ and $\epsilon$=$1$e$-8$ for total 10,000 epochs. The optimization is conducted with total batch size 4 on 2 NVIDIA A30 GPUs, whose learning rate is initially set to $2$e$-4$ and gradually decays to $1$e$-5$ following a cosine attenuation schedule. During training, we augment the samples by random cropping, flipping, rotating and reversing channel order to prevent overfitting. Proposed window partition cropping method is also employed with input resolution of 512$\times$512 and 128$\times$128 for the two-stage SAFNet. For better efficiency, optical flow and selection masks are predicted at 1/2 input resolution and then upsampled back. The overall training takes less than 24 hours.

\noindent \textbf{Comparison on Kalantari 17 dataset.} We compare proposed algorithm with 11 well-known multi-exposure HDR methods on Kalantari 17 test dataset~\cite{Kalantari_2017_ToG}, whose quantitative results are summarized in Table~\ref{tab:2}. As can be seen, proposed SAFNet achieves state-of-the-art performance on all five metrics consistently. It is worth noting that our method surpasses the second-best result~\cite{9881970} by 0.58 dB and 0.0015 in terms of PSNR-$l$ and SSIM-$l$, because of our explicit HDR fusion process in linear domain as Eq.~\ref{eq:5}. Figure~\ref{fig:1} and Figure~\ref{fig:6} qualitatively compare SAFNet with 6 famous HDR algorithms. It is obvious that our method can generate more realistic background with much fewer ghosting artifacts.

To compare execution efficiency of different solutions, we download their official code and evaluate the running time, model parameters and computation complexity on a machine equipped with one NVIDIA A30 GPU under 1500$\times$1000 resolution. Inference time is averaged over 100 iterations. In Table~\ref{tab:2}, FlexHDR~\cite{9881970}, HDR-Transformer~\cite{Liu_2022_ECCV} and SCTNet~\cite{Tel_2023_ICCV} achieve similar PSNR-$\mu$ accuracy as ours. However, running speed of SAFNet is about \textbf{10}$\times$, \textbf{18}$\times$ and \textbf{23}$\times$ faster than them, respectively. We attribute the excellent efficiency of SAFNet to its coarse-to-fine fully convolutional deep architecture without any complex attention mechanism. In regard to computation complexity, proposed SAFNet saves about 37\%, 48\% and 55\% multiply-add operations than advanced SCTNet~\cite{Tel_2023_ICCV}, HDR-Transformer~\cite{Liu_2022_ECCV} and AHDRNet~\cite{Yan_2019_CVPR}, separately. For comparison with the efficient NHDRRNet~\cite{8989959} which also leverages the pyramidal encoder-decoder architecture, we build a small SAFNet-S by reducing feature channels of the refine module from 80 to 32 and replacing residual blocks with convolutions as shown in Figure~\ref{fig:3}. As listed in Table~\ref{tab:2}, SAFNet-S has similar running time as NHDRRNet~\cite{8989959}, but obtains significant performance improvement. In conclusion, proposed SAFNet sets new records of speed-accuracy trade-off on multi-exposure HDR imaging task, which is depicted in Figure~\ref{fig:1} intuitively.

\begin{figure*}[t]
	\begin{minipage}{0.60\linewidth}
		\begin{table}[H]
			\renewcommand{\arraystretch}{0.6}
			{\scriptsize
				\centering
				\setlength\tabcolsep{0.1pt}
				\caption{Quantitative comparison on Challenge123 test set. The middle and bottom parts are trained on patches of 256$\times$256 and 512$\times$512, respectively.}
				\begin{tabular}{l|ccccc}
					\toprule
					Method & PSNR-$\mu$ & PSNR-$l$ & SSIM-$\mu$ & SSIM-$l$ & HDR-VDP2 \\
					\midrule
					Sen~\cite{Sen_2012_ToG} & 37.11 & 27.80 & 0.9729 & 0.9687 & 51.93 \\
					Kalantari~\cite{Kalantari_2017_ToG} & 37.83 & 29.62 & 0.9707 & 0.9705 & 51.32 \\
					\midrule
					AHDRNet~\cite{Yan_2019_CVPR} & 40.44 & 28.13 & 0.9877 & 0.9703 & 54.58 \\
					NHDRRNet~\cite{8989959} & 37.82 & 26.75 & 0.9769 & 0.9632 & 53.38 \\
					HDR-Tran.~\cite{Liu_2022_ECCV} & 40.70 & 28.72 & 0.9881 & 0.9731 & 54.63 \\
					SCTNet~\cite{Tel_2023_ICCV} & 40.65 & 28.73 & 0.9882 & 0.9721 & 54.35 \\
					SAFNet (Ours) & 40.94 & 28.93 & 0.9885 & 0.9740 & 54.84 \\
					\midrule
					AHDRNet~\cite{Yan_2019_CVPR} & 40.61 & 28.33 & 0.9880 & 0.9708 & 54.97 \\
					NHDRRNet~\cite{8989959} & 37.44 & 26.31 & 0.9762 & 0.9596 & 53.51 \\
					SAFNet (Ours) & \textbf{41.88} & \textbf{29.73} & \textbf{0.9897} & \textbf{0.9784} & \textbf{55.07} \\
					\bottomrule
				\end{tabular}
				\label{tab:3}}
		\end{table}
	\end{minipage}
	\hspace{0.00\linewidth}
	\begin{minipage}{0.38\linewidth}
		\begin{figure}[H]
			\tiny
			\centering
			\includegraphics[width=1.00\textwidth]{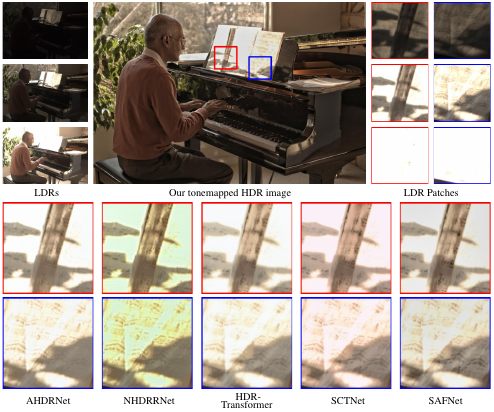}
			\caption{Generalization comparison on the Sen's dataset~\cite{Sen_2012_ToG}.}
			\label{fig:7}
		\end{figure}
	\end{minipage}
\end{figure*}

\noindent \textbf{Evaluation on Challenge123 dataset.} For evaluation on the developed Challenge123 dataset, we retrain 4 advanced methods~\cite{Yan_2019_CVPR,8989959,Liu_2022_ECCV,Tel_2023_ICCV} on our training set with their official implementations, and also compare two alignment-based algorithms~\cite{Sen_2012_ToG,Kalantari_2017_ToG}. Due to the 24GB memory limitation for NVIDIA A30 GPU, transformer-based methods~\cite{Liu_2022_ECCV,Tel_2023_ICCV} can only be optimized up to 256$\times$256 resolution. Therefore, we first train all approaches on 256$\times$256 patches under the same data augmentation method and learning schedule for fair comparison. After training, we summarize the accuracy of different algorithms on our test set in the middle part of Table~\ref{tab:3}. It can be concluded that our SAFNet consistently performs well on all five metrics, exceeding HDR-Transformer~\cite{Liu_2022_ECCV}, SCTNet~\cite{Tel_2023_ICCV}, AHDRNet~\cite{Yan_2019_CVPR} and NHDRRNet~\cite{8989959} by 0.24, 0.29, 0.50 and 3.12 dB on PSNR-$\mu$, separately, verifying our superior multi-exposure HDR architecture.

Since our dataset contains larger motion magnitude than Kalantari 17~\cite{Kalantari_2017_ToG}, training crop size of AHDRNet~\cite{Yan_2019_CVPR}, NHDRRNet~\cite{8989959} and our SAFNet are extended to 512$\times$512 for further comparison on long-range texture aggregation. In the bottom of Table~\ref{tab:3}, our SAFNet still outperforms all the others on all metrics, and our performance advantages are further amplified in challenging cases. Increasing patch size from 256 to 512, proposed SAFNet improves most, AHDRNet~\cite{Yan_2019_CVPR} improves smaller than ours, while NHDRRNet~\cite{8989959} even drops in some metrics. It indicates that our SAFNet is more robust and capable to learn from challenging large motion training samples. Transformer-based solutions~\cite{Liu_2022_ECCV,Tel_2023_ICCV} still fall behind our SAFNet, since their patch-based prediction manner can not aggregate cross-patch moving texture in high-resolution photography. Furthermore, in the top of Table~\ref{tab:3}, alignment-based HDR fusion approach~\cite{Kalantari_2017_ToG} behaves well in PSNR-$l$, confirming the significance of alignment for large motion. Differently, our SAFNet exceeds~\cite{Kalantari_2017_ToG} on all metrics, due to our stronger registration and fusion ability. More analysis can be found in our supplementary.

Figure~\ref{fig:6} visually compares them on a challenging sunset scene, including both camera and clouds motion. We can observe that early flow and patch based methods~\cite{Sen_2012_ToG,Kalantari_2017_ToG} generate distorted edges since misalignment in saturated regions. Attention-based networks~\cite{Yan_2019_CVPR,8989959} synthesize blurry outputs due to lack of motion compensation. Transformer-based solutions~\cite{Liu_2022_ECCV,Tel_2023_ICCV} suffer from block artifacts because of limited aggregation scope. In contrast, our SAFNet can fuse more pleasing HDR images. More results can be found in our supplementary.

\begin{table}[b]
	\centering
	\renewcommand{\arraystretch}{0.5}
	{\scriptsize
		\centering
		\setlength\tabcolsep{7.4pt}
		\caption{Ablation on first fusion stage. $F_{2\rightarrow i}$ and $M_i$ are output components of the decoder. Accuracy is measured on $H_m$.}
		\begin{tabular}{cc|cccc}
			\toprule
			$F_{2\rightarrow i}$ & $M_i$ & PSNR-$\mu$ & PSNR-$l$ & SSIM-$\mu$ & SSIM-$l$ \\
			\midrule
			\cmark & \xmark & 33.69 & 36.30 & 0.9568 & 0.9701 \\
			\xmark & \cmark & 40.69 & 37.08 & 0.9834 & 0.9772 \\
			\cmark & \cmark & \textbf{41.68} & \textbf{39.61} & \textbf{0.9851} & \textbf{0.9808} \\
			\bottomrule
		\end{tabular}
		\label{tab:4}}
\end{table}

\noindent \textbf{Generalization Ability.} To compare generalization capability of recent HDR imaging methods, we test 5 algorithms on the unsupervised Sen's dataset~\cite{Sen_2012_ToG}, where all methods are trained on our proposed Challenge123 dataset fairly. As shown in Figure~\ref{fig:7}, NHDRRNet~\cite{8989959} and SCTNet~\cite{Tel_2023_ICCV} suffer from unnatural color problem, while AHDRNet~\cite{Yan_2019_CVPR} and HDR-Transformer~\cite{Liu_2022_ECCV} synthesize more hazy textures compared to proposed SAFNet, especially in the shaded areas.

\begin{table}[t]
	\centering
	\renewcommand{\arraystretch}{0.5}
	{\scriptsize
		\centering
		\setlength\tabcolsep{5.3pt}
		\caption{Ablation on second refinement stage. $F_{2\rightarrow i}$, $M_i$ and $H_m$ are input of the refine network. Accuracy is measured on $H_r$.}
		\begin{tabular}{ccc|cccc}
			\toprule
			$F_{2\rightarrow i}$ & $M_i$ & $H_m$ & PSNR-$\mu$ & PSNR-$l$ & SSIM-$\mu$ & SSIM-$l$ \\
			\midrule
			\xmark & \xmark & \xmark & 43.63 & 41.67 & 0.9922 & 0.9902 \\
			\cmark & \xmark & \xmark & 43.73 & 41.88 & 0.9924 & 0.9902 \\
			\cmark & \cmark & \xmark & 43.87 & 42.04 & 0.9925 & 0.9904 \\
			\cmark & \cmark & \cmark & \textbf{44.59} & \textbf{43.15} & \textbf{0.9929} & \textbf{0.9911} \\
			\bottomrule
		\end{tabular}
		\label{tab:5}}
\end{table}

\begin{table}[t]
	\centering
	\renewcommand{\arraystretch}{0.5}
	{\scriptsize
		\centering
		\setlength\tabcolsep{4.8pt}
		\caption{Ablation on window partition cropping. S1 and S2 are input resolution for two stages. Accuracy is measured on $H_r$.}
		\begin{tabular}{cc|cccc}
			\toprule
			S1 & S2 & PSNR-$\mu$ & PSNR-$l$ & SSIM-$\mu$ & SSIM-$l$ \\
			\midrule
			128$\times$128 & 128$\times$128 & 44.59 & 43.15 & 0.9929 & 0.9911 \\
			512$\times$512 & 512$\times$512 & 44.54 & 43.09 & 0.9930 & 0.9914 \\
			512$\times$512 & 128$\times$128 & \textbf{44.66} & \textbf{43.18} & \textbf{0.9932} & \textbf{0.9917} \\
			\bottomrule
		\end{tabular}
		\label{tab:6}}
\end{table}

\begin{figure}[t]
	\tiny
	\centering
	\includegraphics[width=0.48\textwidth]{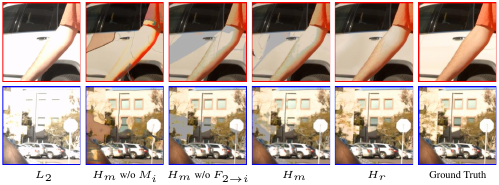}
	\caption{Qualitative results of ablation on selection mask, optical flow and refinement.}
	\label{fig:8}
\end{figure}

\begin{figure}[t]
	\tiny
	\centering
	\resizebox{0.5\textwidth}{!}{
		\begin{tabular}{@{}c @{\hskip 0.01in} c @{\hskip 0.01in} c @{\hskip 0.01in} c @{\hskip 0.01in} c@{}}
			\includegraphics[width=0.168\linewidth]{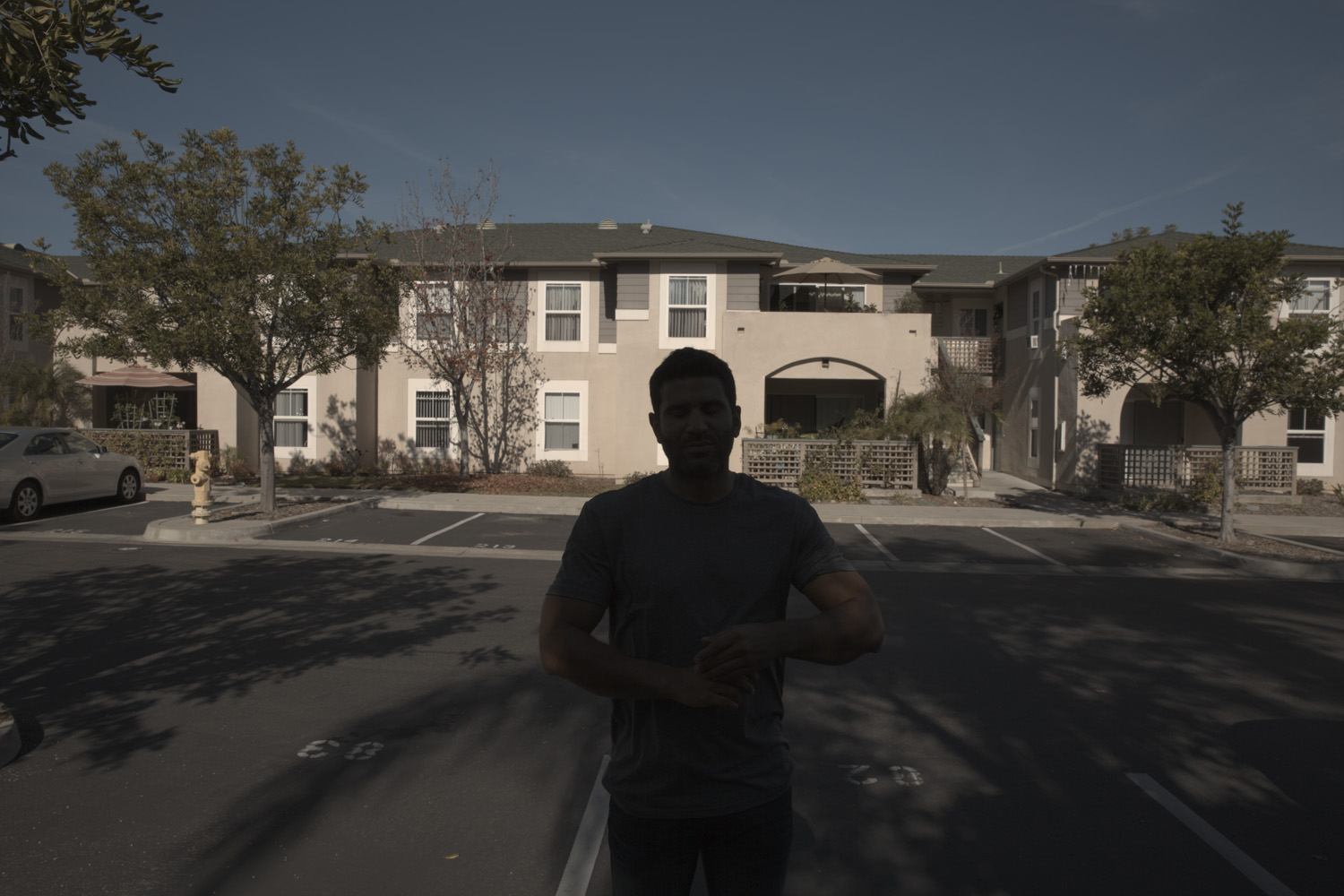}
			&
			\includegraphics[width=0.168\linewidth]{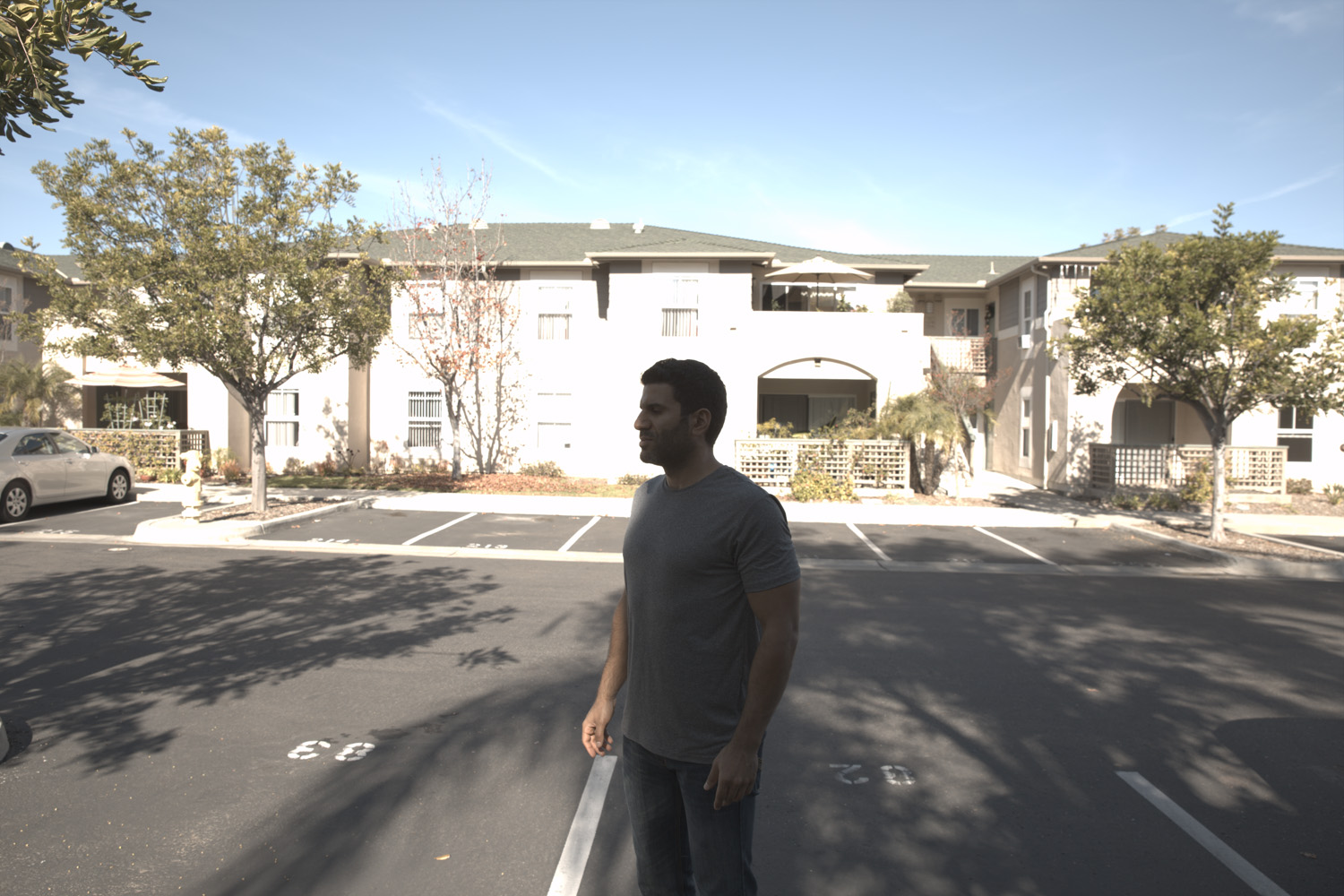}
			&
			\includegraphics[width=0.168\linewidth]{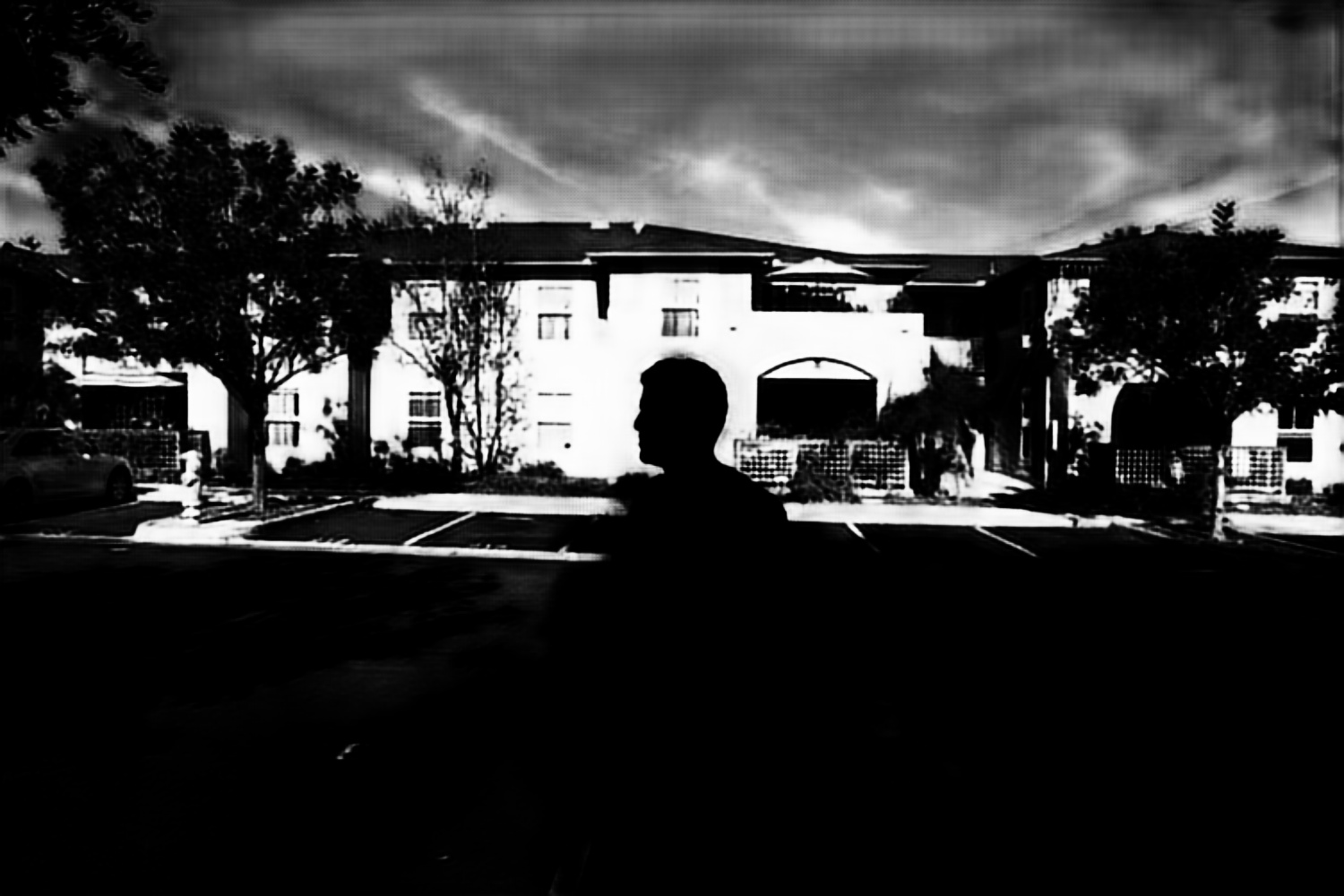}
			&
			\includegraphics[width=0.168\linewidth]{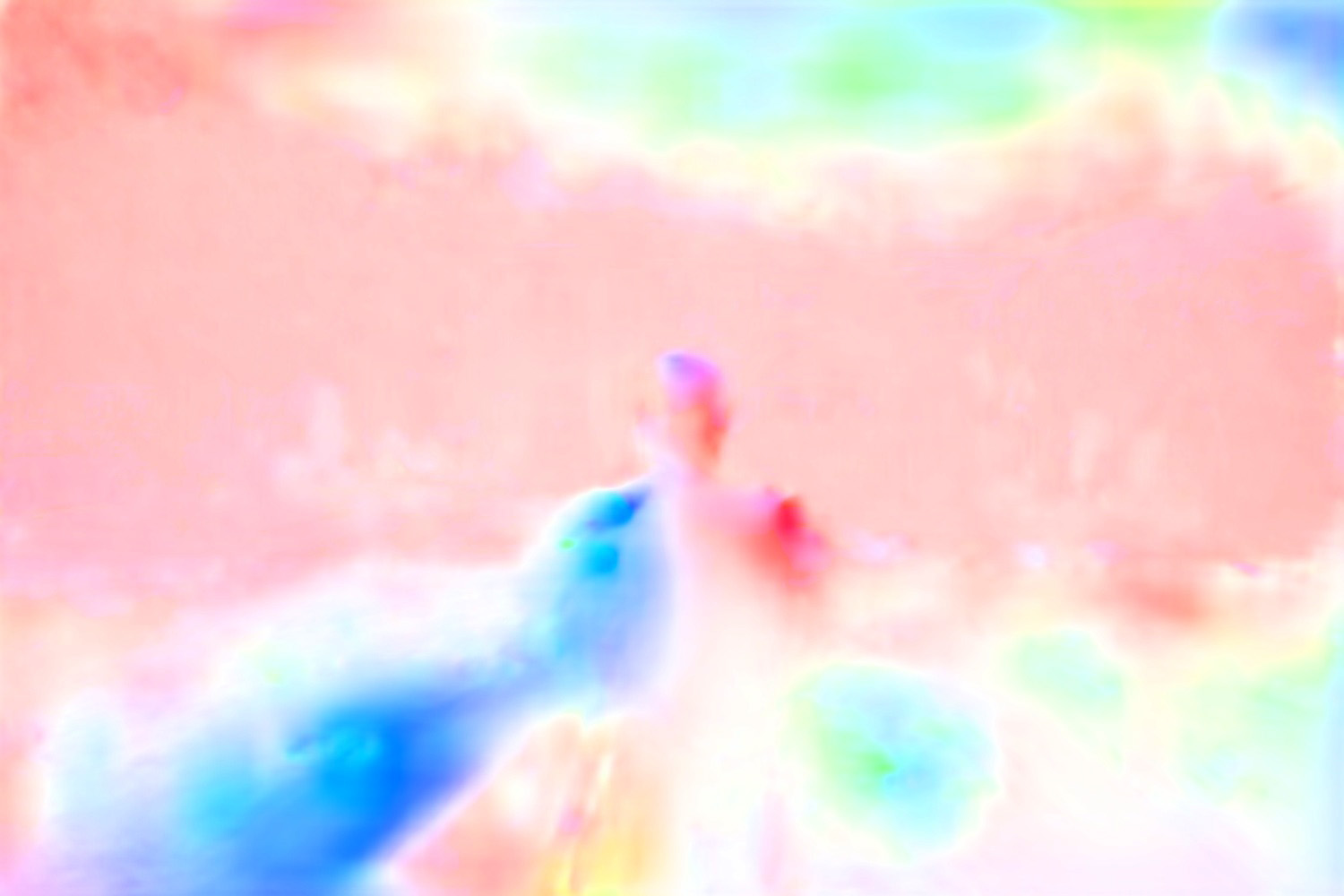}
			\\
			\includegraphics[width=0.168\linewidth]{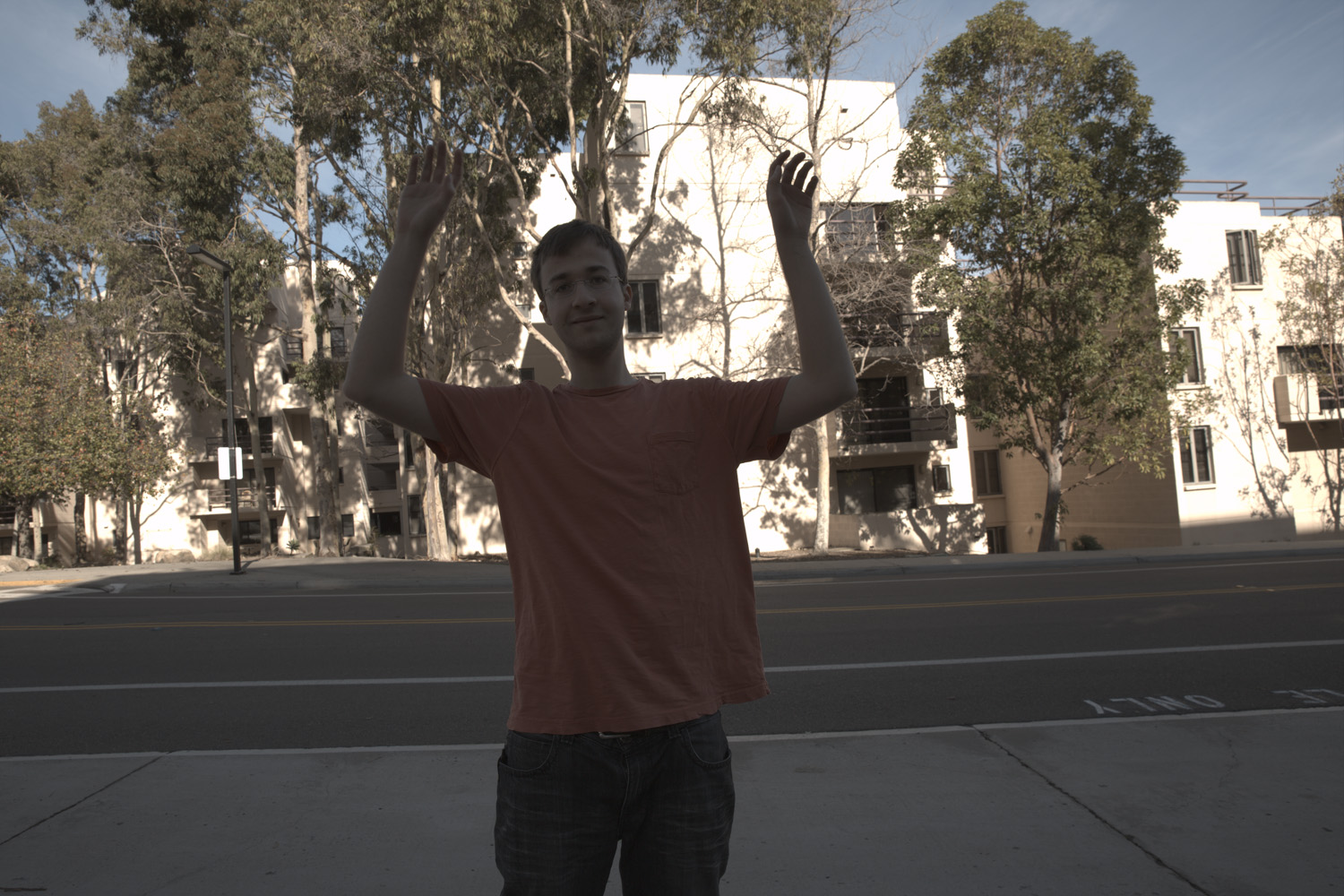}
			&
			\includegraphics[width=0.168\linewidth]{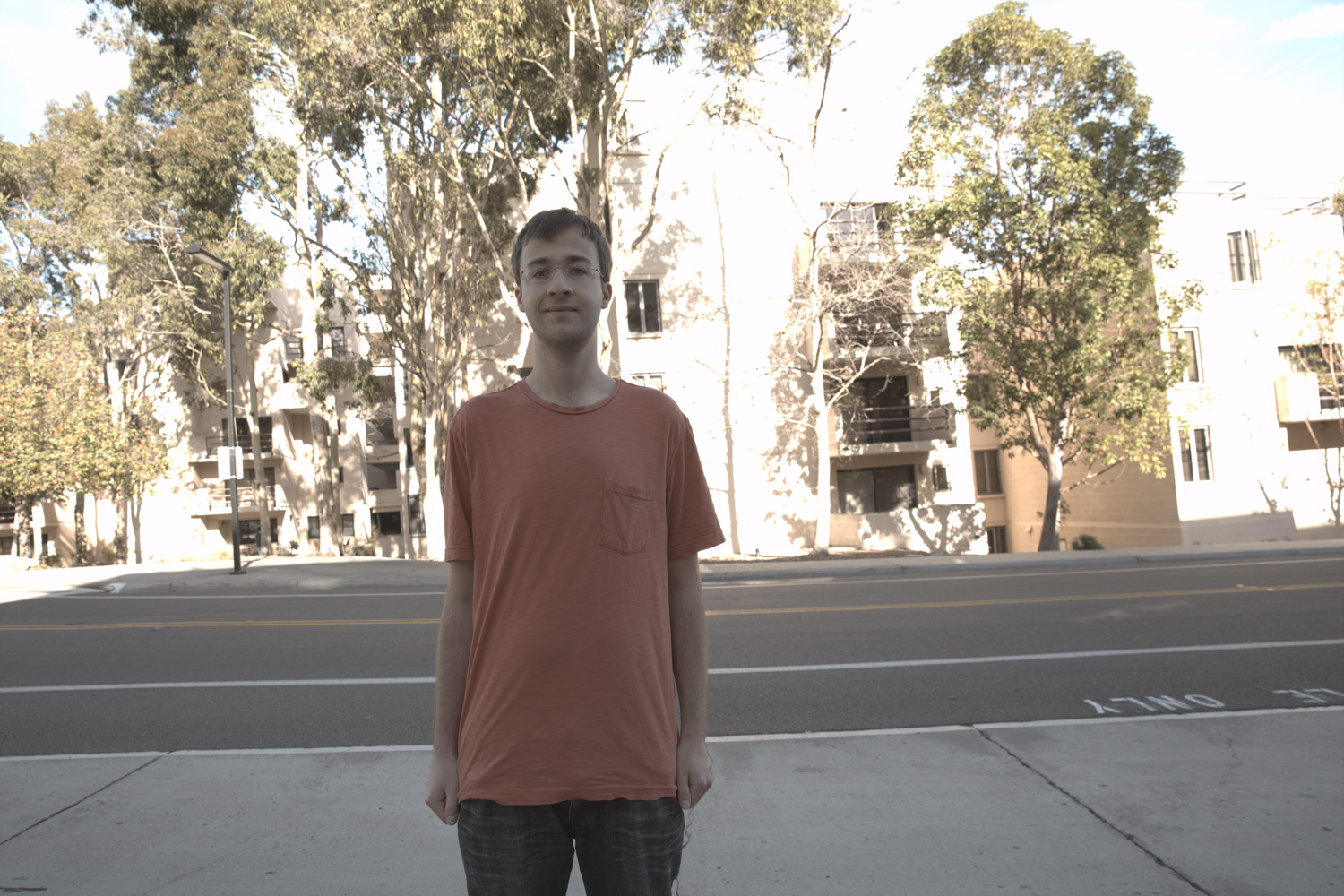}
			&
			\includegraphics[width=0.168\linewidth]{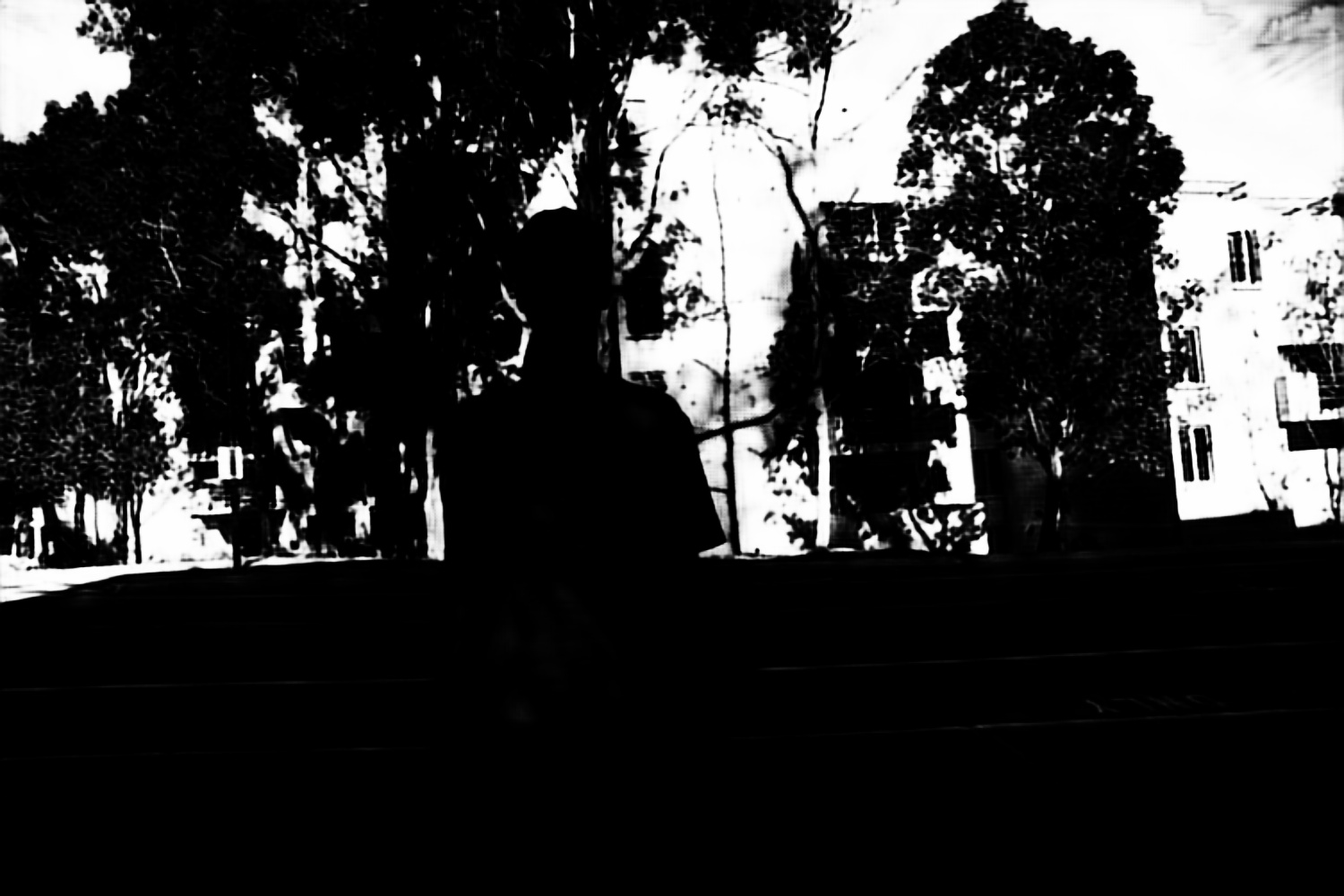}
			&
			\includegraphics[width=0.168\linewidth]{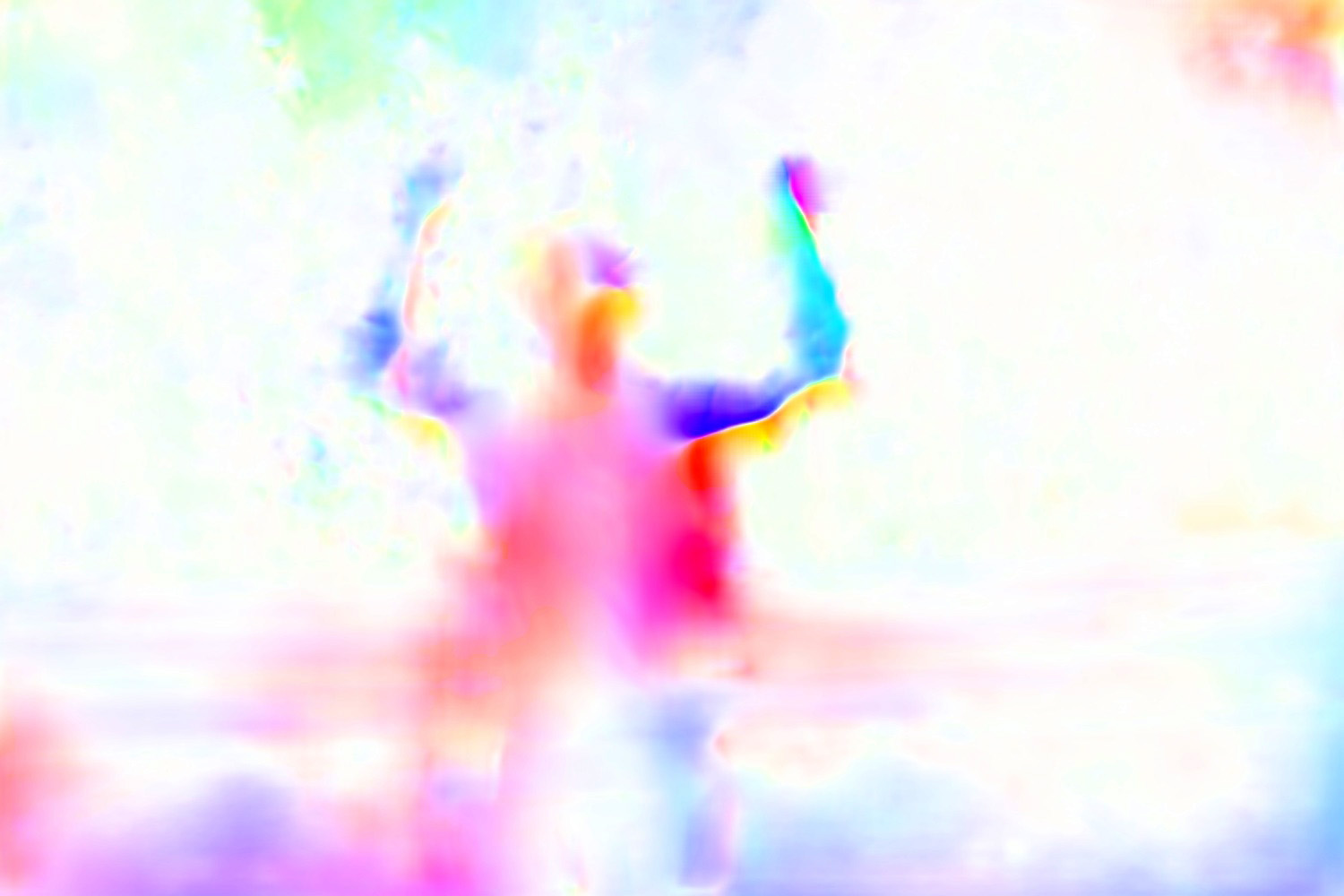}
			\\
			$L_1$ & $L_2$ & $M_1$ & $F_{2\rightarrow1}$
			\\
	\end{tabular}}
	\caption{Visualization of selection mask $M_1$ and optical flow $F_{2\rightarrow1}$.}
	\label{fig:9}
\end{figure}

\noindent \textbf{Ablation on First Fusion Stage.} In the first stage, SAFNet explicitly merges an initial HDR image $H_m$ as in Eq.~\ref{eq:5}. To verify the effectiveness of optical flow $F_{2\rightarrow i}$ and selection masks $M_i$ predicted by the decoder, we do ablations on Kalantari 17 dataset~\cite{Kalantari_2017_ToG} by selectively removing them in the progressive refinement procedure. In Table~\ref{tab:4}, removing selection masks $M_i$ will result in significant performance degradation, since precise alignment for entire image is extremely challenging under heavy saturation and complex motion. On the other hand, removing optical flow $F_{2\rightarrow i}$ will cause a smaller but also noticeable accuracy loss, because of lost ability for long-range texture aggregation. Figure~\ref{fig:8} visually compares these ablation experiments. As can be seen, $H_m$ w/o $M_i$ looks more twisted, while $H_m$ w/o $F_{2\rightarrow i}$ lacks texture in moving regions. By jointly refining selection masks and optical flow in selected regions, proposed approach can concentrate on finding and fusing more valuable textures for HDR reconstruction more efficiently. Figure~\ref{fig:9} depicts two visual examples of selection masks and optical flow predicted by SAFNet. We can observe that our network can find over-exposed wall in the reference LDR image and estimate relatively precise optical flow in selected regions to aggregate valuable cross-exposure textures. As for unselected black regions, flow estimation of SAFNet is relatively free without worry about negative impacts, such as potential ghosting artifacts.

\noindent \textbf{Ablation on Second Refinement Stage.} In the second stage, SAFNet compensates high frequency details by the refine network based on LDR inputs and first stage outputs, \textit{i.e.}, optical flow $F_{2\rightarrow i}$, selection masks $M_i$ and merged HDR image $H_m$. To explore the value of different inputs for the refinement module, we gradually remove $H_m$, $M_i$ and $F_{2\rightarrow i}$ from the inputs, and carry out ablations on Kalantari 17 dataset~\cite{Kalantari_2017_ToG}. As listed in Table~\ref{tab:5}, optical flow, selection masks and merged HDR image all contain valuable while complementary information for guiding the refinement procedure, where the best result is achieved when using all these input components. It is worth noting that $H_m$ contributes most to the performance, confirming the significance of region selective HDR fusion in the first stage. Figure~\ref{fig:8} depicts two visual examples to demonstrate the effectiveness of our refine subnetwork, which can not only compensate for missing contextual details but also correct distorted scene structure.

\noindent \textbf{Ablation on Window Partition Cropping.} To confirm the effectiveness of our window partition cropping method, we do ablations on different input resolutions when optimizing SAFNet, whose results are summarized in Table~\ref{tab:6}. It can be seen that when employing proposed window partition cropping approach to train SAFNet on different input resolutions, we can yield the best performance. We attribute the reason to that there is trade-off between training patch size and HDR accuracy. Larger patch size can create more long-range aggregation training samples for merging $H_m$, while smaller input patches can generate more challenging samples with diverse occlusion and saturation cases for refining $H_r$.

\section{Conclusion}
\label{sec:conclusion}
In this paper, we present a novel SAFNet for efficient and accurate multi-exposure HDR imaging. By jointly refining valuable area selection masks and optical flow in selected regions, it can focus on finding and aggregating more useful LDR textures and finally merge a high quality HDR image explicitly. Based on diverse features exported from the first fusion stage, a lightweight refinement module is introduced to compensate for missing details. Moreover, to better optimize our two-stage SAFNet, a new window partition cropping method is proposed. Experiments on conventional and newly developed challenging datasets demonstrate that our algorithm not only outperforms recent SOTA methods quantitatively and qualitatively, but also runs order of magnitude faster than advanced transformer-based solutions.

\title{SAFNet: Selective Alignment Fusion Network for Efficient HDR Imaging - Supplementary Material}

\titlerunning{SAFNet: Selective Alignment Fusion Network for Efficient HDR Imaging}

\author{Lingtong Kong \and
%\orcidlink{0000-0003-2212-3581}
Bo Li \and
Yike Xiong \and
Hao Zhang \and
Hong Gu \and
Jinwei Chen\textsuperscript{\Letter}}

\authorrunning{L. Kong et al.}

\institute{vivo Mobile Communication Co., Ltd, China\\
\email{\{ltkong,libra,cokexiong,haozhang,guhong,jinwei.chen\}@vivo.com}}

\maketitle

\section{Challenge123 Dataset}
The existing labeled multi-exposure HDR datasets~\cite{Froehlich_2014,Kalantari_2017_ToG,liu2023mobilehdr,Tel_2023_ICCV} have facilitated research in related fields. Nevertheless, results of recent methods~\cite{Liu_2022_ECCV,9881970,Yan_2023_CVPR,Tel_2023_ICCV} tend to be saturated due to their limited evaluative ability~\cite{Kalantari_2017_ToG,Tel_2023_ICCV}. We attribute this phenomenon to their relatively small motion magnitude between LDR inputs and relatively small saturation ratio of the reference LDR image. To widen the performance gap between different algorithms, we propose a new challenging multi-exposure HDR dataset with enhanced motion range and saturated regions, whose statistics comparison with existing datasets~\cite{Kalantari_2017_ToG,Tel_2023_ICCV} is listed in Table 1 of our main paper. In the supplementary material, we elaborate construction details of the proposed Challenge123 HDR dataset.

\begin{figure*}[t]
	\centering
	\includegraphics[width=0.98\linewidth]{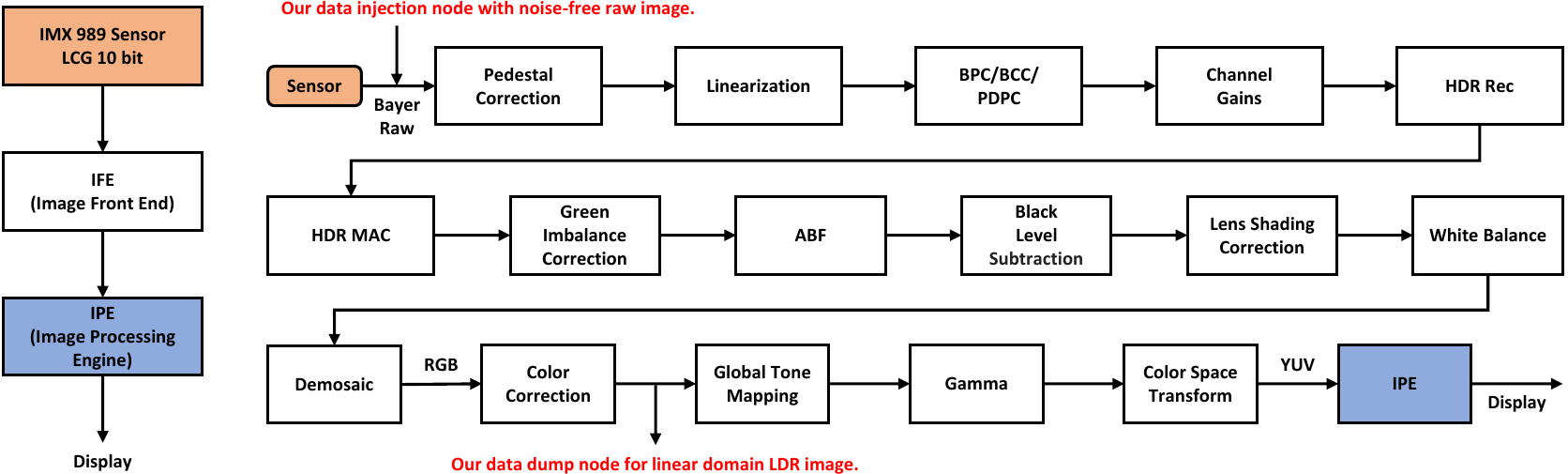}
	\caption{Details of our ISP simulation pipeline. The left part shows overall framework of the ISP pipeline for Qualcomm platform. The right part presents details of the Image Front End (IFE) in Qualcomm platform. We inject Bayer raw data before `Pedestal Correction' and dump simulated LDR image before `Global Tone Mapping'.}
	\label{fig:10}
\end{figure*}

To capture LDR raw image, we use a vivo X90 Pro+ phone equipped with high-end Sony IMX 989 sensor under different lighting conditions, containing indoor, outdoor, daytime and nighttime scenarios. To obtain LDR images with different exposures in a controlled dynamic scene, we first make the camera to automatically adjust exposure, white balance and focus parameters according to the integrated algorithms of the smart phone for better adaptation. Secondly, we fix all camera parameters except for the shutter speed to capture LDR sequences with three different exposures, \textit{i.e.}, under-, middle- and over-exposure. Our mobile phone is fixed on a tripod to keep steady, and we take 10 to 100 successive frames per exposure for subsequent denoising. Generally, the number of shots grows when exposure time decreases, varying based on different noise levels. Finally, we can obtain the noise-free LDR raw image for each exposure by averaging all successive raw frames under the same camera parameter.

To acquire HDR ground truth, we first copy above noise-free LDR raw images with their corresponding camera parameter files from the mobile phone to a desktop computer. Then, we perform a relatively comprehensive ISP simulation pipeline to generate high quality LDR images in the linear domain, whose implementation details are depicted in Figure~\ref{fig:10}. Specifically, we use the parameter parser and simulation tool provided by Qualcomm, which can parse the camera parameter `.bin' file into `.xml' file, and simulate a relatively complete Image Front End (IFE) pipeline based on `.raw' and `.xml' files, respectively. We dump the intermediate result before Global Tone Mapping (GTM) in RGB color space as our simulated LDR image in linear domain. Finally, we merge above linear domain LDR images of three different exposures by using the weighting function in \cite{Debevec_1997,Kalantari_2017_ToG}, generating high quality HDR ground truth and reference LDR image. As for the non-reference LDR images, we follow the same acquisition process as before, but move the mobile phone with a relatively large camera pose to create relatively large inter-frame motion. Also, exposure time of the reference LDR image is set to a larger or a smaller value than the normal one for generating more saturated regions. The above two approaches can make our paired LDR-HDR dataset more challenging than the existing ones~\cite{Kalantari_2017_ToG,Tel_2023_ICCV}, which has been analyzed in Table 1 and Table 3 of our main paper.

Based on above data collection and processing strategy, we develop a labeled multi-exposure HDR dataset, called Challenge123 dataset, including 96 training samples and 27 test samples, covering diverse lighting conditions, shooting time, motion modes and scene structures. To enhance the applicability of our dataset and promote future research, for each of three content-related moving scenes, we further create under-, middle- and over-exposure LDR images and corresponding HDR image. It means that for each of our 96 training scenes, we have $3 \times 2 \times 1 = 6$ exposure combination for training theoretically, while all experiments on our Challenge123 dataset in this paper adopt under-, middle- and over-exposure LDR images by the time order like previous methods.

\section{More Results and Analysis}
\begin{figure*}[t]
	\centering
	\begin{minipage}{0.48\linewidth}
		\begin{figure}[H]
			\includegraphics[width=0.97\columnwidth]{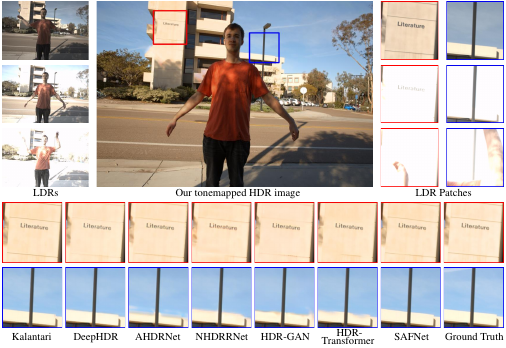}
			\caption{Visual Comparison on Kalantari 17 test dataset~\cite{Kalantari_2017_ToG}. Zoom in for best view.}
			\label{fig:11}
		\end{figure}
	\end{minipage}
	\hspace{0.00\linewidth}
	\begin{minipage}{0.48\linewidth}
		\begin{figure}[H]
			\includegraphics[width=0.97\columnwidth]{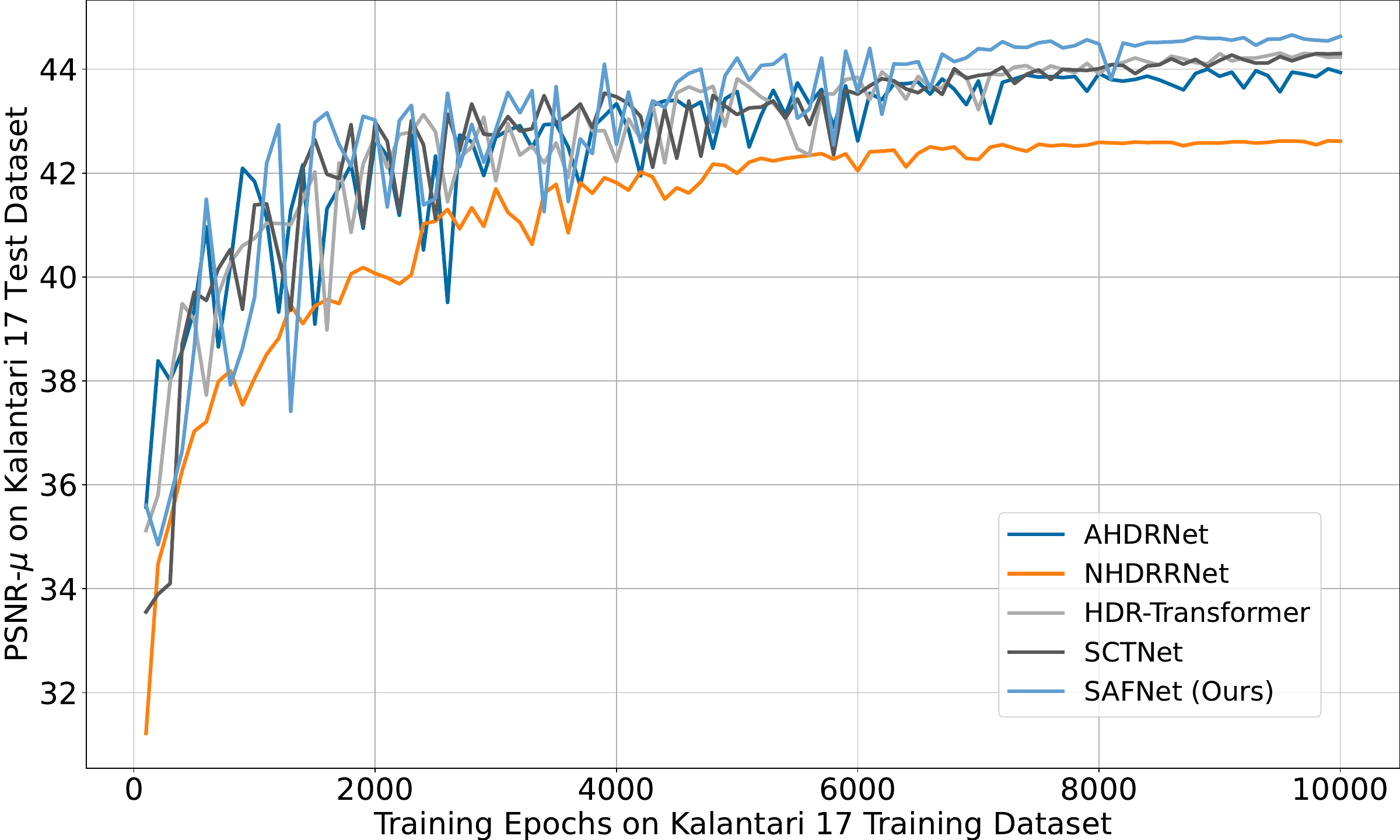}
			\caption{PSNR-$\mu$ on Kalantari 17 benchmark~\cite{Kalantari_2017_ToG} during the whole training process. All algorithms are compared fairly.}
			\label{fig:12}
		\end{figure}
	\end{minipage}
\end{figure*}

\noindent \textbf{Results on Kalantari 17 Dataset.} In Figure~\ref{fig:11}, we present one more visual comparison on Kalantari 17 test dataset~\cite{Kalantari_2017_ToG}, which compares on non-rigid foreground motion and rigid background motion areas. It is obvious that our SAFNet can not only deal with complex motion and occlusion cases like attention-based methods~\cite{Yan_2019_CVPR,8989959,9387148,Liu_2022_ECCV}, but also generate faithful scene structures as alignment-based approaches~\cite{Kalantari_2017_ToG,Wu_2018_ECCV}.

To fairly compare recent SOTA methods~\cite{Yan_2019_CVPR,8989959,Liu_2022_ECCV,Tel_2023_ICCV} with proposed SAFNet on the well-known Kalantari 17 dataset~\cite{Kalantari_2017_ToG}, we further train all these algorithms under the same data augmentation approach and learning schedule, whose results are summarized in Figure~\ref{fig:12}. As can be seen, proposed SAFNet exceeds AHDRNet~\cite{Yan_2019_CVPR}, NHDRRNet~\cite{8989959}, HDR-Transformer~\cite{Liu_2022_ECCV} and SCTNet~\cite{Tel_2023_ICCV} on PSNR-$\mu$ consistently in the convergence stage. We attribute the reason to that attention-based multi-exposure HDR methods show relatively poor generalization ability on new dynamic scenes, which tend to overfit on existing training samples. Differently, our SAFNet based on explicit flow alignment generalize well on unobserved dynamic scenes. Besides, our joint refinement decoder can adaptively adjust fusion weights according to flow uncertainty in current location, that can merge high quality HDR image with much fewer ghosting artifacts.

\begin{figure}[H]
	\includegraphics[width=0.98\columnwidth]{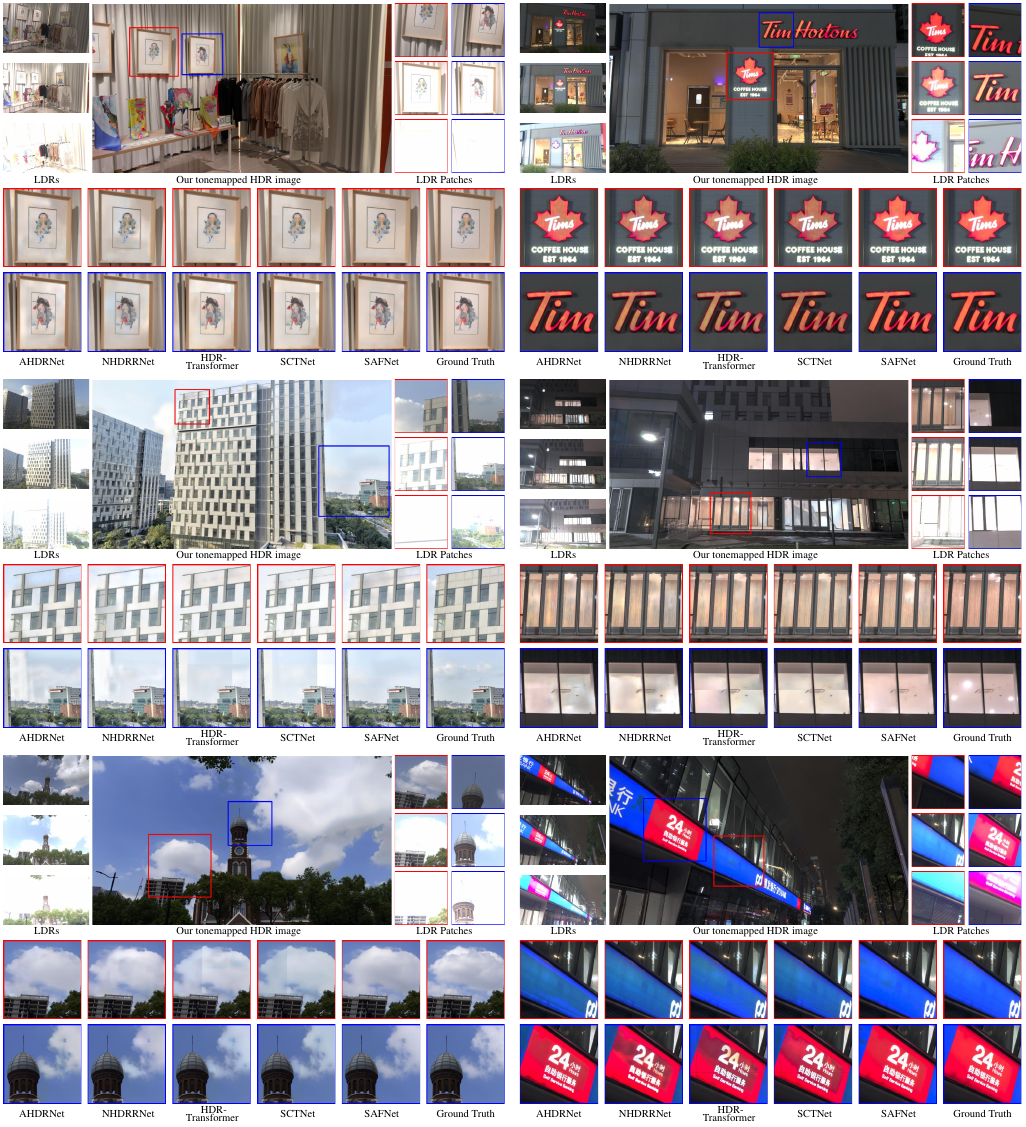}
	\caption{Visual comparison of recent SOTA methods on our Challenge123 test dataset.}
	\label{fig:13}
\end{figure}

\noindent \textbf{Results on Challenge123 Dataset.} In Figure~\ref{fig:13}, we show six more visual comparisons on our developed Challenge123 test dataset, including both daytime and nighttime scenarios. In the top left figure, there are less artifacts on the murals of our SAFNet prediction. In the top right figure, logos and texts reconstructed by our method are more distinct and realistic. In the middle left figure, the building edge and the sky are more faithful regarding to proposed algorithm.  In the middle right figure, texture details synthesized by proposed approach are more natural and coherent. In the bottom left figure, clouds generated by our SAFNet are more consistent and true-colored. In the bottom right figure, signs and texts predicted by our network look more comfortable even if the reference frame is ill-exposed. Summarily, proposed SAFNet can generate more favorable HDR images in challenging motion and exposure scenes.

Note that our proposed Challenge123 dataset aims to widen the performance gap between different algorithms for ease of analysis. The similar result can also be observed on Kalantari 17 Dataset~\cite{Kalantari_2017_ToG}. For example, the first result of HDR-Transformer~\cite{Liu_2022_ECCV} in Figure 6 of our main paper contains block artifacts at the door handle. With the increasing popularity of high-resolution photography, offsets of several hundred pixels are more common in multi-frame HDR imaging, especially in bracket exposure or night scene modes with long time-lapses.

\noindent \textbf{Results on Tel 23 Dataset.} To verify the effectiveness of proposed approaches in more motion types and light conditions, we further train our SAFNet on Tel 23~\cite{Tel_2023_ICCV} training set from scratch with the same learning schedule as on our Challenge123 dataset. Then, we evaluate our algorithm on Tel 23 test set, and compare the results in Table~\ref{tab:7}. It can be observed that our algorithm achieves best accuracy on PSNR-$l$, SSIM-$\mu$ and SSIM-$l$, but falls behind HDR-Transformer~\cite{Liu_2022_ECCV} and SCTNet~\cite{Tel_2023_ICCV} on PSNR-$\mu$ and HDR-VDP2. We attribute the reason to different motion and saturation characteristics between Tel 23 and Kalantari 17 datasets. Kalantari 17 and our developed Challenge123 datasets both contain regions that are both saturated and moving in the reference LDR image, which can evaluate not only the long-range texture aggregation ability but also the deghosting ability of multi-exposure HDR algorithms. Differently, backgrounds of Tel 23 dataset are almost static, while the moving people are nearly well-exposed in the reference LDR frame. Therefore, Tel 23 dataset can only evaluate the deghosting ability when merging multiple LDR images.

\begin{table}[t]
	\centering
	\renewcommand{\arraystretch}{0.5}
	{\scriptsize
		\centering
		\setlength\tabcolsep{5.0pt}
		\caption{Quantitative comparison on Tel 23 test set~\cite{Tel_2023_ICCV}. The best result is in \textbf{bold}.}
		\begin{tabular}{l|ccccc}
			\toprule
			Method & PSNR-$\mu$ & PSNR-$l$ & SSIM-$\mu$ & SSIM-$l$ & HDR-VDP2 \\
			\midrule
			Sen~\cite{Sen_2012_ToG} & 39.97 & 44.21 & 0.9792 & 0.9932 & 67.20 \\
			Kalantari~\cite{Kalantari_2017_ToG} & 41.67 & 47.33 & 0.9838 & 0.9961 & 72.25 \\
			AHDRNet~\cite{Yan_2019_CVPR} & 44.16 & 50.29 & 0.9896 & 0.9971 & 78.12 \\
			HDR-Transformer~\cite{Liu_2022_ECCV} & 44.88 & 51.09 & 0.9904 & 0.9981 & 78.87 \\
			SCTNet~\cite{Tel_2023_ICCV} & \textbf{44.93} & 51.73 & 0.9906 & 0.9981 & \textbf{79.53} \\
			SAFNet (Ours) & 44.61 & \textbf{51.97} & \textbf{0.9908} & \textbf{0.9982} & 78.80 \\
			\bottomrule
		\end{tabular}
		\label{tab:7}}
\end{table}

In conclusion, Transformer-based methods~\cite{Liu_2022_ECCV,Tel_2023_ICCV} are better to deal with multi-exposure image fusion for deghosting. Differently, our flow-based SAFNet are better to handle the moving texture aggregation in a region selective way.

\bibliographystyle{splncs04}
\bibliography{main}
\end{document}